\documentclass[10pt,twocolumn,letterpaper]{article}

\usepackage{iccv}              

\def\methodName{MulSMo}

\definecolor{iccvblue}{rgb}{0.21,0.49,0.74}
\usepackage[pagebackref,breaklinks,colorlinks,allcolors=iccvblue]{hyperref}
\usepackage{colortbl}
\usepackage{multirow}
\usepackage{float}
\usepackage{tabularx}
\usepackage{algorithm}
\usepackage{algpseudocode}
\usepackage{cases}
\usepackage{soul}
\usepackage{wrapfig}
\usepackage{graphicx}
\definecolor{color1}{RGB}{255,250,205}
\definecolor{color2}{RGB}{255,228,225}
\definecolor{color3}{RGB}{34,139,34}

\definecolor{hku_color}{HTML}{13A983} 
\definecolor{szu_color}{HTML}{84193E} 
\definecolor{color4}{RGB}{218,112,214}

\def\paperID{2340} 
\def\confName{ICCV}
\def\confYear{2025}

\title{MulSMo: Multimodal Stylized Motion Generation by Bidirectional Control Flow}
\author{
Zhe Li$^{1 *}$,
Yisheng He$^{2}$,
Zhong Lei$^{3}$,
Weichao Shen$^{2}$,
Qi Zuo$^{2}$,
Lingteng Qiu$^{2}$,\\
~Shenhao Zhu$^{2}$,
Zilong Dong$^{2}$,
Laurence T. Yang$^{1 \dagger}$,
Weihao Yuan$^{2 \dagger}$ \\
$^{1}$ Huazhong University of Science and Technology \\
$^{2}$ Alibaba Group \\
$^{3}$ The University of Edinburgh \\
{ \tt\small ${*}$\quad keycharon0122@gmail.com \qquad $\dagger$\quad Corresponding Authors}
}

\usepackage{amsmath,amsfonts,bm}

\def\eqref#1{equation~\ref{#1}}

\def\1{\bm{1}}

\def\vc{{\bm{c}}}

\def\vg{{\bm{g}}}

\def\vr{{\bm{r}}}
\def\vs{{\bm{s}}}

\def\vx{{\bm{x}}}

\def\vz{{\bm{z}}}

\DeclareMathAlphabet{\mathsfit}{\encodingdefault}{\sfdefault}{m}{sl}
\SetMathAlphabet{\mathsfit}{bold}{\encodingdefault}{\sfdefault}{bx}{n}

\begin{document}
\maketitle

\begin{abstract}

Generating motion sequences conforming to a target style while adhering to the given content prompts requires accommodating both the content and style.
In existing methods, the information usually only flows from style to content, which may cause conflict between the style and content, harming the integration.
Differently, in this work we build a bidirectional control flow between the style and the content, also adjusting the style towards the content, in which case the style-content collision is alleviated and the dynamics of the style is better preserved in the integration.
Moreover, we extend the stylized motion generation from one modality, i.e. the style motion, to multiple modalities including texts and images through contrastive learning, leading to flexible style control on the motion generation.
To further boost the performance, we advance the motion diffusion to motion-aligned temporal latent diffusion by developing a novel motion VAE.
Extensive experiments demonstrate that our method significantly outperforms previous methods across different datasets, while also enabling multimodal signals control.
The code of our method will be made publicly available.

\end{abstract}


\section{Introduction}

Stylized human motion generation is a rapidly growing field in computer vision, offering significant value across various applications like filmmaking, gaming, virtual reality, and robotics.
The model is tasked to generate human motions according to instruction texts while conforming to a given style, which requires the model to accommodate both the content and style features.

Previous methods tackling this problem try to directly concatenate the two features of content and style~\cite{qian2024smcd}, or integrate them by cross attention~\cite{kim2024most}.
Recent methods borrow the ideas from stylized image generation, employing AdaIn~\cite{huang2017arbitrary}, LoRA~\cite{hu2021lora}, or ControlNet~\cite{zhang2023adding} to control the style of the generated motion.
AdaIn is an effective mechanism for enforcing global styles, yet it inherently has limitations when it comes to capturing local details of styles.
LoRA has been successfully utilized to incorporate the target style into the generated content, but it requires training a separate LoRA model for each style, which is laborious for plentiful styles.
In contrast, ControlNet exploits only one model to settle various detailed styles of the same mode, by taking in the control signal of the target style.
Therefore, in this paper, we seek to push the limit of the ControlNet for stylized motion generation by designing a novel approach.


\begin{figure}[]
\centering
 \includegraphics[width=1.0\columnwidth, trim={0cm 0cm 0cm 0cm}, clip]{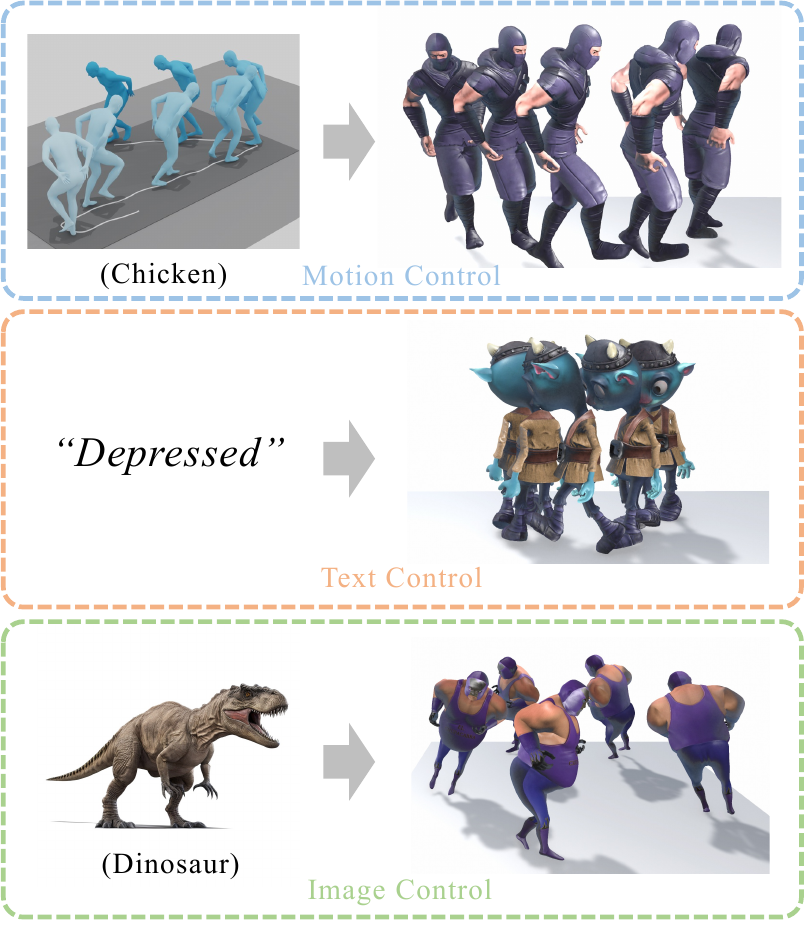}
\vspace{-6mm}
\caption{\textbf{MulSMo} enables multimodal signals to control the stylized motion generation.}
\label{fig:teaser}
\vspace{-6mm}
\end{figure}

\begin{figure*}[t]
\centering
  \includegraphics[width=1.8\columnwidth, trim={0cm 0cm 0cm 0cm}, clip]{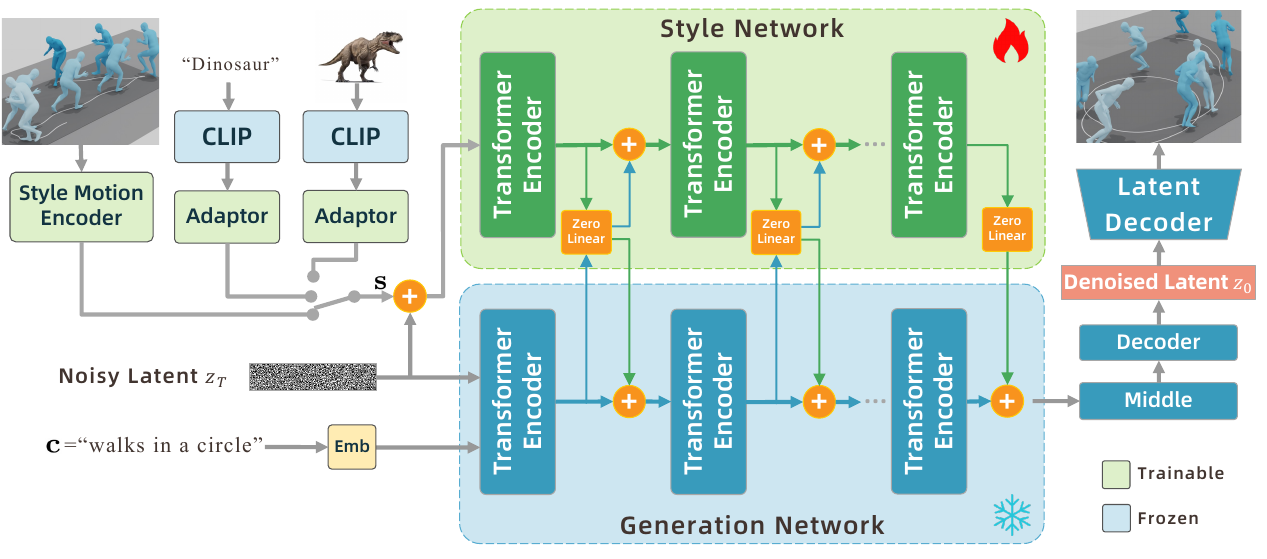}
\vspace{-4mm}
\caption{Overview of MulSMo. Our approach can take in various modalities as style signals, such as style motion sequences, texts, and images. We generate stylized human motions by combining content text $\mathbf{c}$ with style signals $\mathbf{s}$. In the encoder part of the latent denoiser, we establish a bidirectional control flow between the generation network and the style network, utilizing respective zero linear for fusion. (Note that for simplicity we plot the style-to-content zero linear and content-to-style zero linear into one block in the figure.)
This denoising step is repeated $T$ times to obtain the noise-free motion latent $\vz_0$, which is decoded to stylized motion by the latent decoder.
}
\label{fig:framework}
\vspace{-4mm}
\end{figure*}

While previous methods have achieved impressive results by adapting ControlNet for stylized motion generation~\cite{zhong2025smoodi}, the control flow always moves from the style motion to the content motion.
This brings a problem. The style motion often diverges from the content motion, leading to conflict in the integration of style and motion, such that the generation model tends to ignore the conflicted feature in the style.
The conflict is especially severe in the motion dynamics, making it difficult to accurately transfer the dynamics of the style motion to the content motion.
Therefore, the styles differing in dynamics would generate similar motions, such as `aeroplane' and `flapping'.
Another problem of previous methods lies in the control signals, which are limited to a style motion in previous methods~\cite{jang2022motion,guo2024generative,song2024arbitrary,zhong2025smoodi}.
This leads to difficulties for users to input the target style, limiting the applications of stylized motion generation.

To address these two problems, in this paper, we propose a \textbf{Mul}timodal \textbf{S}tylized \textbf{Mo}tion generation model (\methodName), where a bidirectional control mechanism is designed to impose the style onto content motion, and the contrastive learning is leveraged to enable multi-modality style control, as shown in Figure~\ref{fig:teaser}. 
Specifically, we establish a bidirectional control flow between the style network and generation network, such that the control not only flows from the style to content, but also flows from the content to style, as displayed in Figure~\ref{fig:framework}.
This enables the style to also adjust towards the content, bridging the gap between two spaces and relieving the conflict, especially the divergence in dynamics.
To enable both local and global bidirectional control, we equip the bidirectional flow modules in each encoder block so that local features can be controlled in the shallow layers and more global features is controlled while going deeper. 
To enable multimodal signals for style control, we propose to align the style embeddings across style motion sequences, text descriptions, and images within a contrastive learning manner. In this way, the style text descriptions or style images can play the same role as style motion sequences for style control during generation.

To further boost the performance, we also advance the latent diffusion to motion-aligned temporal latent diffusion (MaTLD) by designing a Motion-aligned Temporal VAE, which employs a temporal gating mechanism to enhance sequential pattern preservation and aligns the motion embedding towards motion priors.
Extensive experiments are performed with either style motion, texts, or images as style input, and demonstrate that our method obtains better performance compared to previous methods.

The main contributions of this work are then summarized as follows:
\begin{itemize}

    \item We novelly build a bidirectional motion ControlNet for stylized motion generation, where the control flow moves not only from the style to content but also from the content to style, in which case the style-content collision is relieved.

    \item We extend the stylized motion generation from controlled by only style motion sequence to multimodal signals, where the style embeddings of different-modality signals are aligned by contrastive learning.

    \item We advance the diffusion generation to motion-aligned temporal latent diffusion, which leads to a substantial enhancement in the motion generation.

    \item We outperform previous stylized motion generation and motion style transfer on multiple datasets, while enabling flexible style input from motions, texts, or images.

\end{itemize}

\section{Related Work}

\subsection{Stylized Image Generation}
Image style is generally represented by the global statistical characteristics of images.
Early work by Gatys et al~\cite{gatys2016image} demonstrates the feasibility of transferring visual styles through aligning Gram matrices within neural networks.
Adaptive instance normalization (AdaIN)~\cite{huang2017arbitrary} is introduced to facilitate arbitrary image style transfer by dynamically adapting feature statistics.
%
Recently, diffusion models~\cite{rombach2022high} have achieved great success in image generation, and ControlNet~\cite{zhang2023adding} and LoRA~\cite{hu2021lora} are widely work~\cite{jones2024customizing} used to control these models to generate stylized images.
Specifically, ZipLoRA~\cite{shah2025ziplora} allows for the generation of any subject in any style by effectively combining independently trained style and subject LoRAs.
StyleAdapter~\cite{wang2024styleadapter} introduces a unified stylized image generation model capable of producing a variety of stylized images without the need for test-time fine-tuning.

\subsection{Motion Generation Model}
%
Human motion generation based on transformer models~\cite{li2024lamp, guo2024momask} and diffusion models~\cite{dai2024motionlcm,chen2024motionclr,andreou2024lead} has made significant progress in recent years.
Momask~\cite{guo2024momask} employs a motion residual VQ-VAE with multiple codebooks, enhancing generation through a residual transformer. LaMP~\cite{li2024lamp} designs a novel motion-aware text encoder instead of CLIP to enhance the motion generation quality and proposes a motion-to-text large language model. MDM~\cite{tevet2023human} establishes a solid baseline for text-to-motion generation and serves as a robust pretrained model for novel conditional motion generation.
MLD~\cite{chen2023executing} further enhances efficiency by embedding the diffusion process within the latent space.
Inspired by the efficacy of diffusion models for control and conditioning, several works have controlled pretrained motion diffusion models to follow constraints such as trajectories~\cite{xie2023omnicontrol,karunratanakul2023gmd,wan2023tlcontrol}, 3D scenes~\cite{yi2025generating,cen2024generating}, and object interactions~\cite{peng2023hoi,wu2024thor,christen2024diffh2o}.

\subsection{Stylized Motion Generation}
Motion style transfer~\cite{aberman2020unpaired,park2021diverse,jang2022motion,tao2022style,mason2022local,song2024arbitrary,raab2023single,kim2024most,qian2024smcd,tang2023rsmt,wen2021autoregressive,xu2020hierarchical} is the most widely used technique to create stylized motion, transferring the style from a reference motion to a source motion.
Aberman et al.~\cite{aberman2020unpaired} introduce a generative adversarial network to disentangle motion style from content and facilitate their re-composition, eliminating the need for paired data.
Motion Puzzle~\cite{jang2022motion} designs a generative framework that can control the style of individual body parts.
Guo et al.\cite{guo2024generative} utilizes the latent space of pretrained motion models to improve the feature extraction and integration of motion content and style. Song et al.\cite{song2024arbitrary} proposes a diffusion-based approach that integrates trajectory awareness, and achieves style transfer through AdaIN.

However, a major limitation of the above models is that they rely on motion sequences to extract content information, while stylized motion generation~\cite{Ao2023GestureDiffuCLIP,zhong2025smoodi} can generate motions according to given texts while conforming to a target style.
Inspired by the ControlNet, SMooDi~\cite{zhong2025smoodi} enables stylized motion generation from content text descriptions and style motion sequences by introducing two kinds of style guidance to steer the pretrained text-to-motion model toward the target style, realizing fine-grained style control.
Nonetheless, the control flow only moves from the style to the content, leading to severe conflict in some cases.
In contrast, we propose a bidirectional control mechanism to adapt the style towards content, alleviating the conflict.

\begin{figure}[t]
\centering
  \includegraphics[width=1.0\columnwidth, trim={0cm 0cm 0cm 0cm}, clip]{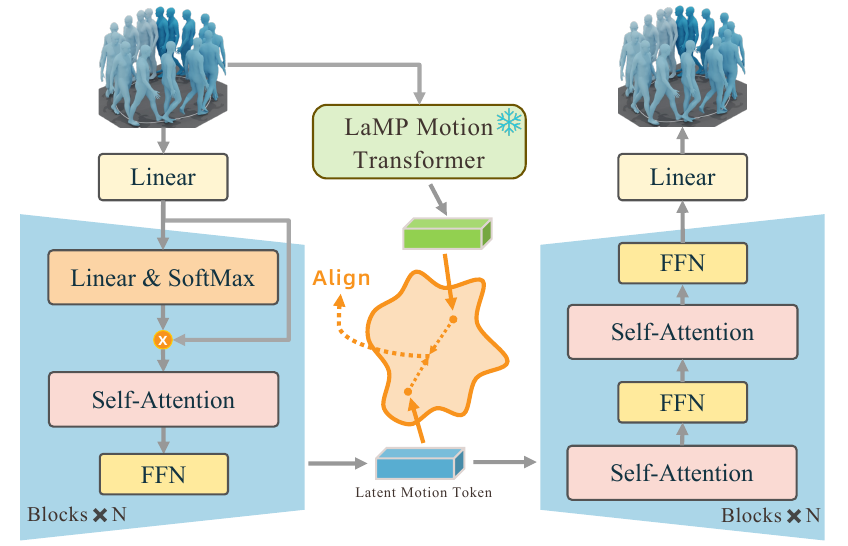}
\vspace{-8mm}
\caption{Motion-aligned Temporal VAE.}
\label{fig:vae}
\vspace{-6mm}
\end{figure}

\section{Method}
\subsection{Overview}
In this section, we present our proposed \methodName, which generates stylized motion sequences by taking text descriptions and multi-modal style signals as input. An overview of our model is depicted in Figure~\ref{fig:framework}. Aligning with MLD \citep{chen2023executing}, we propose MaTLD to operate the diffusion process in the motion latent space with motion-aligned temporal VAE, which is depicted in Figure \ref{fig:vae}. In our method, $\epsilon_{\theta}$ denotes the latent denoiser and $\{\vz_t\}_{t=0}^T$ represents the noised latent sequences, where $\vz_T$ is a Gaussian noise. Our purpose is to denoise from $\vz_T$ to a clean stylized latent motion sequence. Given a content prompt $\mathbf{c}$ and a style signal $\mathbf{s}$, we define the denoising process at step $t \in (0, T]$ as $\epsilon_{t} = \epsilon_{\theta}(\vz_t, t, \mathbf{c}, \mathbf{s})$, where $\epsilon_{\theta}$ denotes the predicted noise at timestep $t$. 
After denoising for $T$ timesteps, we obtain a clean latent $\vz_0$, which is then decoded by a pretrained decoder into a stylized motion sequence $\hat{\vx}_0 \in \mathbb{R}^{L \times D}$, where $L$ denotes the frame length of motion and $D$ is the dimension of human motion representations. Following \citep{zhong2025smoodi}, we adopt the same motion representations as detailed in HumanML3D \citep{Guo_2022_CVPR}, with $D = 263$.
As depicted in Figure~\ref{fig:framework}, the text description $\mathbf{c}$ is used to control the content domain of the generated motion, while the style domain is controlled by style signals, including reference motion sequences, texts, or images. 

\begin{figure}[]
\centering
  \includegraphics[width=0.9\columnwidth, trim={0cm 0cm 0cm 0cm}, clip]{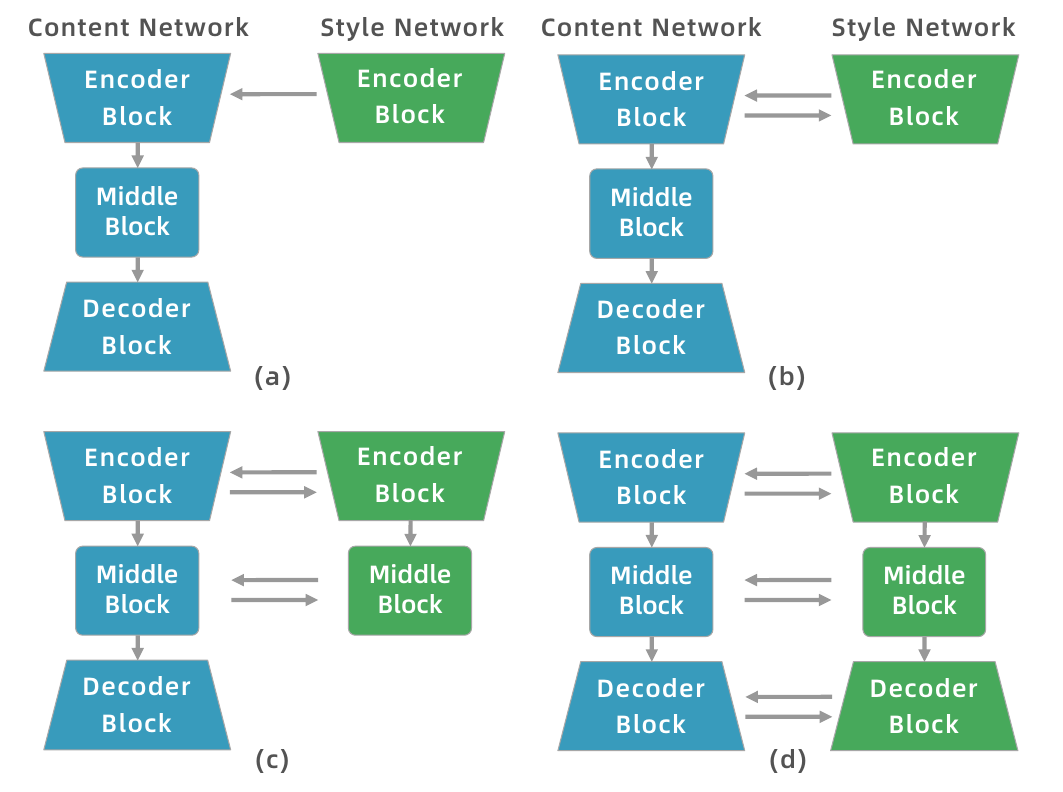}
\vspace{-2mm}
\caption{Architectures with different control flow between generation network and style network. Based on experimental validation, we select (\textbf{b}) as the final architecture.
}
\label{fig:ablation}
\vspace{-5mm}
\end{figure}

\subsection{Motion-aligned Temporal VAE}
In the context of the MLD-based Variational Autoencoder (VAE) \citep{chen2023executing}, the encoder performs resampling on motion sequences of varying lengths to produce a single latent motion token, upon which a diffusion process is conducted. However, this compression along the temporal dimension results in the loss of certain temporal information within the latent token. To mitigate this issue, we employ a gating mechanism to attenuate noise introduced into the attention layers depicted in Figure~\ref{fig:vae}, allowing the model to concentrate more effectively on interactions occurring at critical time steps. Specifically, for each layer of the transformer, prior to the self-attention operation, we apply an MLP to the input features, followed by a softmax operation to compute weights. These weights are then multiplied with the input features to reduce the temporal information loss associated with the resampled single motion token while simultaneously serving to pre-filter noise (e.g., decreasing the weights of frames corresponding to stagnant motion).

Due to the training strategy of the VAE, the latent representations tend to quickly deviate from their initial states in pursuit of reconstruction optimization, which may limit their expressiveness. To address this issue, we align the motion latent tokens with the motion transformer of the LaMP \citep{li2024lamp}, aiming to enhance the expressive capacity of the latent representations while simultaneously regularizing the high-dimensional latent space. Furthermore, diffusion models generate latent vectors through denoising, and their outputs may carry residual noise and a certain degree of stochasticity, which does not align with the clean latent representations expected by the decoder of VAE. Therefore, we posit that this alignment process can also be regarded as an effective means of enhancing the robustness of the decoder. Our proposed motion-aligned temporal VAE is optimized with KL loss, reconstruction loss \citep{kingma2013auto}, and InfoNCE loss \citep{he2020momentum} for alignment:
\begin{subequations}
	\begin{numcases}{} 
	\mathcal{L} = \mathcal{L}_{\text{KL}} + \mathcal{L}_{\text{recon}} + \mathcal{L}_{\text{InfoNCE}} \\
	\mathcal{L}_{\text{InfoNCE}} = -\log \frac{\exp(\text{sim}(t_i, s_j) / \tau_t)}{\sum_{k=1}^{N} \exp(\text{sim}(t_i, s_k) / \tau_t)},
	\end{numcases}
	\label{eq}
\end{subequations}
where $t_i$ and $s_j$ are positive samples, while $t_i$ and $s_k$ are negative samples. sim($\cdot$) means cosine similarity and $\tau$ denotes the temperature hyperparameter.

\begin{figure}[t]
\centering
  \includegraphics[width=0.8\columnwidth, trim={0cm 0cm 0cm 0cm}, clip]{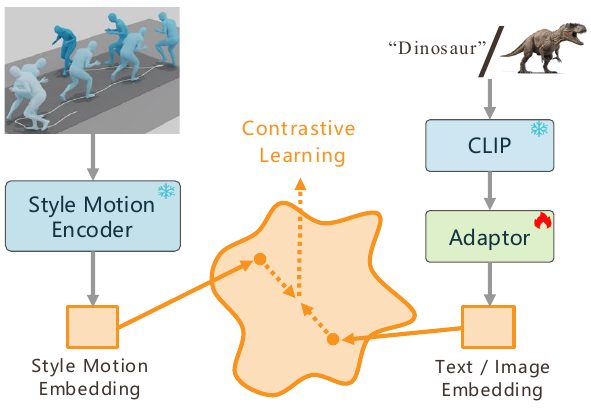}
\vspace{-2mm}
\caption{{Contrastive learning for enabling multi-modality stylized motion control.} The adaptor can partially alleviate the gap between style motions and images/texts.
}
\label{fig:contrastive}
\vspace{-5mm}
\end{figure}

\begin{figure*}[]
\centering
  \includegraphics[width=1.8\columnwidth, trim={0cm 0cm 0cm 0cm}, clip]{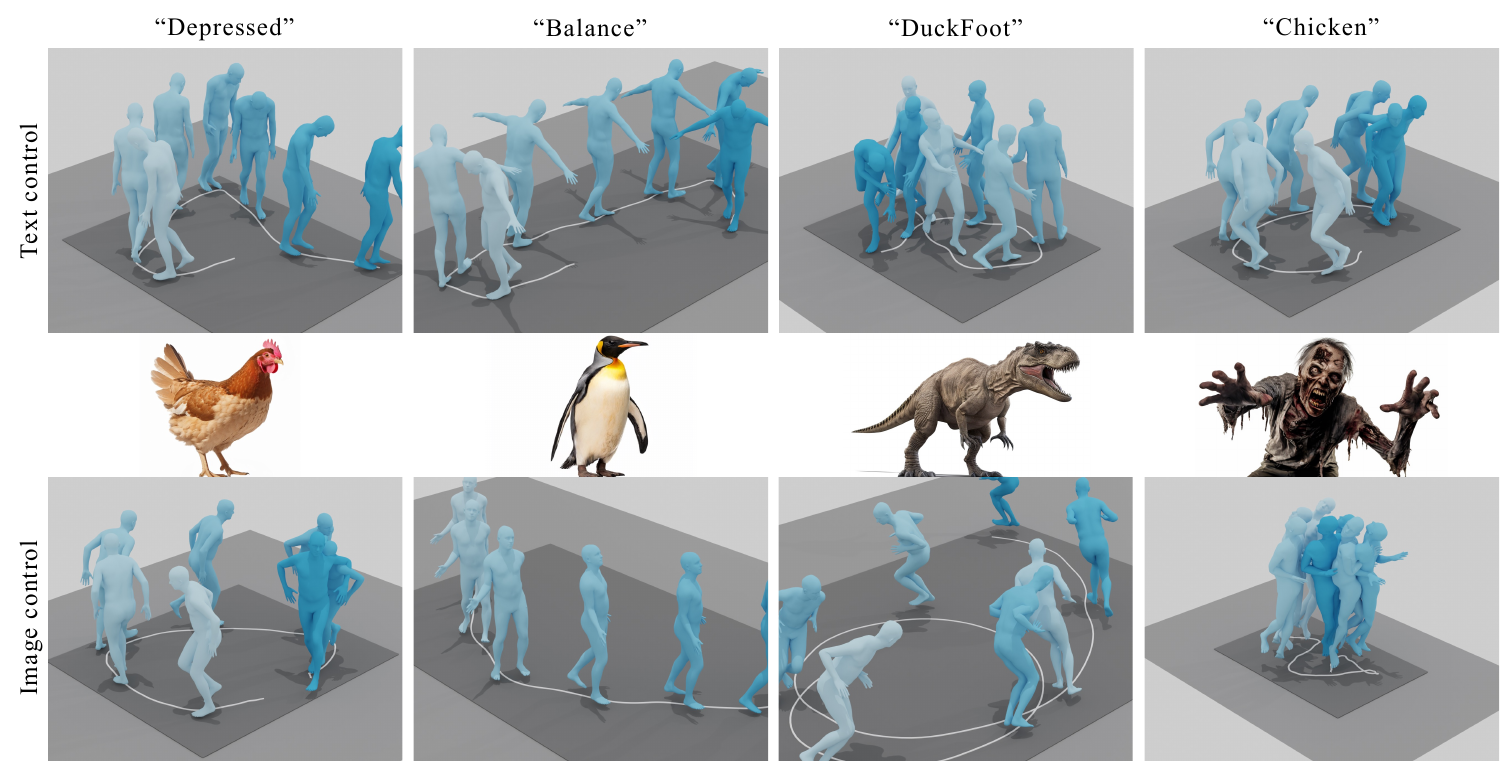}
\vspace{-2mm}
\caption{Qualitative results of using text or image as the style control signal.
}
\label{fig:res_multimodal}
\vspace{-6mm}
\end{figure*}

\subsection{Bidirectional Motion Control Flow}


We improve the ControlNet \citep{zhang2023adding} by introducing a bidirectional feedback mechanism to make the transferred motion style more accurate. As depicted in Figure \ref{fig:ablation}(a), existing works~\cite{zhong2025smoodi} directly inject the control signal into the generation network encoder. The conflict between the style and content features makes the generation unstable. In contrast, as shown in Figure \ref{fig:ablation}(b), we build bidirectional fusion blocks between the style and content encoder to bridge the gap and enable more stable generation. The main idea is that the generation network and the control network engage in a high-frequency interaction, which synchronously adjusts the style and content features in a dynamic manner, minimizing the conflict and maintaining the dynamism of the style feature instead of being a simple pose.  

As illustrated in the middle part of Figure~\ref{fig:framework}, we build bidirectional fusion modules between the style and content encoders to bridge the information gap and resolve the conflict dynamically. To enable both local and global information to be shared between the two encoders, we design bidirectional fusion modules in each encoder block so that local information can be shared in the shallow layers and more global information is exchanged while going deeper. Denote $\mathcal{S} = \{\mathcal{S}_i, i\in\{1,2,...,N\}\}$ the $N$ blocks style encoder, where $\mathcal{S}_i$ is the $i_{th}$ block and denote its output encoded feature as $F_{s_{i}}$. 
Similarly, we denote $\mathcal{G} = \{\mathcal{G}_i, i\in\{1,2,...,N\}\}$ the content encoder, $\mathcal{G}_{i}$ the $i_{th}$ block and $F_{c_{i}}$ the encoded feature of the $i_{th}$ block. Denote fusion modules from style to contents as $\mathcal{L}_{s2c} = \{\mathcal{L}_{s2c_i}, i\in\{1,2,...,N\}\}$. Then the fusion from the style to the content feature at the $i_{th}$ block can be denoted as:
\begin{equation}
    F_{s2c_{i}} = \mathcal{L}_{s2c_i}(F_{s_{i-1}}) + F_{c_{i-1}}.
\end{equation}
Same as the style to content fusion, the content to style fusion modules $\mathcal{L}_{c2s} = \{\mathcal{L}_{c2s_i}, i\in\{1,2,...,N\}\}$ performs fusion at the $i_{th}$ block as:
\begin{equation}
    F_{c2s_{i}} = \mathcal{L}_{c2s_i}(F_{c_{i-1}}) + F_{s_{i-1}}.
\end{equation}
The fused features are then regarded as the output feature of the $i_{th}$ block as $F_{c_i} = F_{s2c_{i}}$ and $F_{s_i} = F_{c2s_{i}}$. They are then fed into the following blocks for further feature extraction and bidirectional fusion in a similar way. In the implementation, we use linear layers to perform fusion modules ($\mathcal{L}_{s2c}$ and $\mathcal{L}_{c2s}$) and initialize them with zero weights following LoRA~\cite{hu2021lora} for stable training.

The designed bidirectional motion control flow bridges the information gap and synchronously resolves the conflict between the style and content. This is crucial for stable stylized motion generation, especially when the content and the style are in different domains, e.g., using texts or images for style control, which we will discuss in the following section.

\subsection{Multimodal Style Control}
Previous methods, e.g.,~\cite{Ao2023GestureDiffuCLIP,zhong2025smoodi} require style motion sequences for stylized motion control. However, capturing such style motion sequences is time-consuming and restricts these methods for scaling up. In this work, we enable multi-modality stylized motion control so that the generated motion can be easily stylized by text or image control, which is easier to obtain and is more user-friendly. To achieve this, our main idea is to align the feature between the style motion sequences, text descriptions, and the style images by contrastive learning. Specifically, we first utilize CLIP~\cite{radford2021learning} pretrained from numerous image-text pairs to align the features between text descriptions and images. Then the aligned image-text features, after being projected by an adapter, are aligned with the style motion sequence feature within a contrastive learning scheme.

As shown in Figure~\ref{fig:contrastive}, we input the style label into CLIP to obtain text embeddings $t$. After passing through an adaptor, we align the text embeddings $t$ with the style embeddings $s$ and optimize them using the InfoNCE loss. During the inference stage, we can use a single label word as a textual control signal for stylized motion generation. Since CLIP has already aligned language and vision, we can also input images as control signals, which are processed through the adaptor to obtain style-aware image embeddings as style control.

In this way, we alleviate the domain gaps between image, text, and style motion sequence when serving as style control signals. These extracted control features are then fed into the bidirectional motion control flow modules for stylized motion generation. The proposed bidirectional control modules further resolve the conflicts between different domains and enable stable stylized motion generation given only text or image as style control signals.

\begin{figure*}[]
\centering
  \includegraphics[width=2\columnwidth, trim={0cm 0cm 0cm 0cm}, clip]{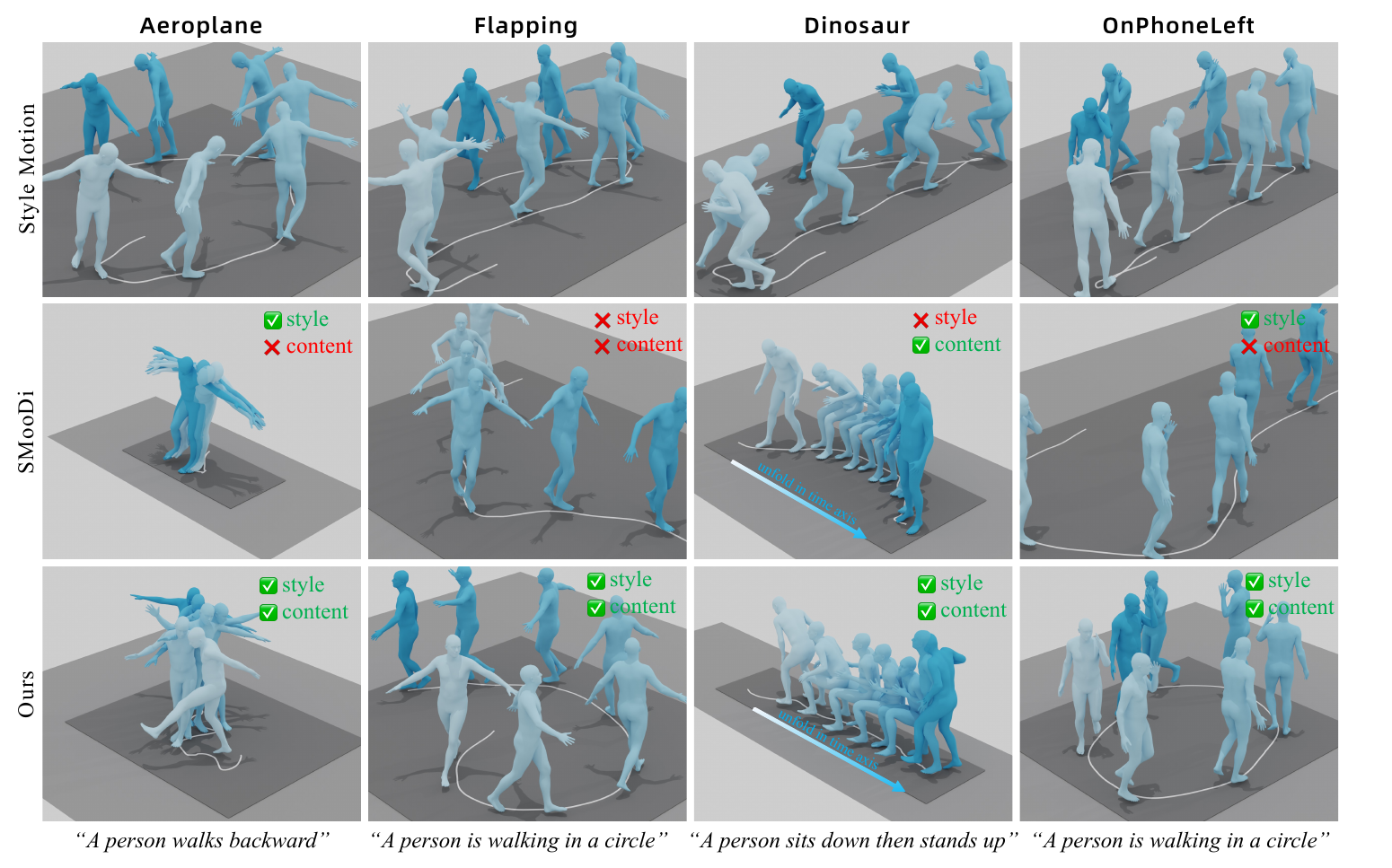}
\vspace{-4mm}
\caption{Qualitative results of stylized motion generation on the 100STYLE dataset. The style input is a style motion while the content input is a text prompt. Note that the `Flapping' style involves the up-and-down motion.
}
\label{fig:res_compare}
\vspace{-5mm}
\end{figure*}

\subsection{Stylized Motion Generation Guidance}
Following \citep{zhong2025smoodi}, we adopt classifier-free guidance and classifier guidance to further enhance the stylized motion generation process. The combination of these two guidance methods makes the generated motion more consistent with the content prompt and the style signal.

\noindent \textbf{Classifier-free Style Guidance.}
We divide the conditioned classifier-free guidance into two parts.
\begin{align}
\label{eql:cfg}
 \mathbf{\epsilon}_\theta(\vz_t, t, \mathbf{c},\mathbf{s}) = \,& \mathbf{\epsilon}_\theta(\vz_t, t, \emptyset, \emptyset) \notag\\ 
 + &w_c \big(\mathbf{\epsilon}_\theta(\vz_t, t, \mathbf{c}, \emptyset) -  \mathbf{\epsilon}_\theta(\vz_t, t, \emptyset, \emptyset) \big) \notag\\
 + &w_s \big(\mathbf{\epsilon}_\theta(\vz_t, t,\mathbf{c}, \mathbf{s}) -  \mathbf{\epsilon}_\theta(\vz_t, t,\mathbf{c}, \emptyset) \big),
\end{align}
where $w_c$ and $w_s$ represent the strengths of the classifier-free guidance for the condition $\mathbf{c}$ and $\mathbf{s}$, respectively.
We slightly abuse the notations here by using $\emptyset$ to denote a condition is not used.

\noindent \textbf{Classifier-based Style Guidance.}
To further improve the stylization of a motion diffusion model, we adopt the classifier guidance~\cite{dhariwal2021diffusion,yu2023freedom} to provide stronger guidance to the generated motion towards the desired style. 
The core of our classifier-based style guidance is a novel analytic function $G(\vz_t, t, \mathbf{s})$, which calculates the $L_1$ distance between the style embedding of the generated clean motion $\hat{\vx}_0$ at denoising step $t$ and the reference style motions $\mathbf{s}$. 
The function's gradient is utilized to steer the generated motion towards the target style.
\begin{align}
\mathbf{\epsilon}_\theta(\vz_t, t, \mathbf{c}, \mathbf{s}) &= \mathbf{\epsilon}_\theta(\vz_t, t, \mathbf{c},\mathbf{s}) + \tau \nabla_{\vz_t}G(\vz_t, t, \mathbf{s}), \notag\\
G(\vz_t, t, \mathbf{s}) &= |f(\hat{\vx}_0) - f(\mathbf{s})|,
\end{align}
%
where $\tau$ adjusts the strength of reference-style guidance and $f$ denotes the style feature extractor.
The generated motion $\hat{\vx}_0$ is obtained by first converting the denoising output latent $\vz_t$ into the predicted clean latent as shown below:
\begin{equation}
\hat{\vz}_0 =  \frac{\vz_t - \sqrt{1 - \alpha_t} \varepsilon_\theta(\vz_t, t, \vc, \vs)}{\sqrt{\alpha_t}},
\end{equation}
where $\alpha_t$ denotes the pre-defined noise scale in the forward process of the diffusion model. The predicted clean latent $\hat{\vz}_0$ is then input into the motion decoder $\mathbf{D}$ to obtain the generated motion.

The style feature extractor is obtained by training a style classifier on the 100STYLE dataset~\cite{mason2022local} and removing its final layer. Training the style feature extractor with ground-truth style labels for supervision enables it to effectively capture style-related features. Consequently, style classifier guidance can provide more precise support for stylized motion generation.

\setlength{\tabcolsep}{4pt}
\begin{table*}[]
\centering
\begin{tabular}{clcccccc}
\toprule
\small
\multirow{2}{*}{\shortstack{Style\\Condition}} & \multicolumn{1}{c}{\multirow{2}{*}{Method}} & \multirow{2}{*}{SRA(\%)$\uparrow$} & \multirow{2}{*}{FID$\downarrow$}&\multirow{2}{*}{MM Dist$\downarrow$}& R-Precision $\uparrow$ &\multirow{2}{*}{Diversity$\rightarrow$}&Foot Skate\\
&&&&&(Top 3)&&Ratio $\downarrow$\\
\midrule
\multirow{5}{*}{Motion}
&MLD+Aberman \citep{aberman2020unpaired}&54.367&3.309&5.983&0.406&8.816&0.347\\
&MLD+Motion Puzzle \citep{jang2022motion}&63.769&6.127&6.467&0.290&6.476&0.185\\
&SMooDi \citep{zhong2025smoodi}&72.418&1.609&4.477&0.571&\cellcolor{color1}9.235&0.124\\
&BiFlow + MLD&\cellcolor{color1} 77.042&\cellcolor{color1}1.527&\cellcolor{color1}4.292&\cellcolor{color1}0.613&\cellcolor{color2}9.303&\cellcolor{color1}0.118\\
&BiFlow + MaTLD (Ours)&\cellcolor{color2}78.029&\cellcolor{color2}1.511&\cellcolor{color2}3.823&\cellcolor{color2}0.676&8.939&\cellcolor{color2}0.116\\
\cline{2-8}
\noalign{\vspace{1pt}}
\multirow{3}{*}{Motion}
&SMooDi$^{*}$ \citep{zhong2025smoodi}&81.298&3.066&5.011&0.439&\cellcolor{color1}{9.433}&0.155\\
&BiFlow + MLD$^{*}$&\cellcolor{color1}86.455&\cellcolor{color1}3.028&\cellcolor{color1}4.872&\cellcolor{color1}0.498&\cellcolor{color2}9.514&\cellcolor{color1}0.153\\
&BiFlow + MaTLD (Ours$^{*}$)&\cellcolor{color2}87.324&\cellcolor{color2}2.977&\cellcolor{color2}4.636&\cellcolor{color2}0.543&8.115&\cellcolor{color2}0.149\\
\midrule
\midrule
\multirow{3}{*}{Text}&ChatGPT+MLD~\cite{zhong2025smoodi} & 4.819 & \cellcolor{color2}0.614 & 4.313 & 0.605 &\cellcolor{color1} 8.836 & 0.131\\
&Adaptor + MLD& \cellcolor{color1}52.100 & 0.754 & \cellcolor{color1}3.875 &\cellcolor{color1} 0.674 & \cellcolor{color2}8.875& \cellcolor{color2}0.104\\
&Adaptor + MaTLD (Ours)& \cellcolor{color2}52.307 & \cellcolor{color1}0.749 & \cellcolor{color2}3.587 &\cellcolor{color2}0.710& 8.687 &\cellcolor{color1} 0.114\\
\bottomrule
\end{tabular}
\vspace{-2mm}
\caption{Quantitative comparison with baseline methods of stylized motion generation on 100STYLE dataset. We utilize style motion and text as control signals respectively. 
Note that there is a tradeoff between style reflection metrics (SRA) and content preservation metrics (R-precision, FID, MM Dist).
SMooDi$^*$ and Ours$^*$ indicate that we adjust $w_s$ to improve the style intensity and reach higher SRA.
The best results under the same setting are in bold.
}
\vspace{-4mm}
\label{stylegen}
\end{table*}
\setlength{\tabcolsep}{6pt}

\subsection{Learning Scheme}
We freeze the parameters of MaTLD and only train the style network on the 100STYLE dataset using the following loss function:
\begin{equation}
\mathcal{L}_{std} = \mathbb{E}_{\epsilon, \vz} \left[ \left\| \epsilon_{\theta}(\vz_t, t, \mathbf{c}, \mathbf{s}) - \epsilon \right\|^2_2 \right],
\end{equation}
where $\epsilon \sim \mathcal{N}(0, \mathbf{I})$ represents the ground-truth noise added to $\vz_0$.
Experimentally, when training solely on the 100STYLE dataset, we observe a phenomenon similar to that mentioned in SMooDi~\cite{zhong2025smoodi}: the model gradually tends to resample the reference motions rather than following the given content text.
Therefore, we adopt a content prior preservation loss $L_{pr}$, on the HumanML3D dataset:

\begin{equation}
\mathcal{L}_{pr} = \mathbb{E}_{\epsilon^{\prime},\vz^{\prime}} \left[ \left\| \epsilon_{\theta}(\vz_t^{\prime}, t, \mathbf{c}^{\prime}, \mathbf{s}^{\prime}) - \epsilon^{\prime} \right\|^2_2 \right],
\end{equation}
where $\vz_t^{\prime}$, $\mathbf{c}^{\prime}$ and $\mathbf{s}^{\prime}$ represents the motion latent, content prompt and style motion sequence derived from the HumanML3D dataset.
$\epsilon^{\prime}$ is the noise map added to $\vz^{\prime}_0$.
%
With content prior preservation loss, our model can maintain the ability to follow the content text constraint while learning diverse motion styles from the 100STYLE dataset.

Overall, the training loss function is defined as follows:
\begin{equation}
\mathcal{L}_{all} = \mathcal{L}_{std} + \lambda_{pr} \mathcal{L}_{pr},
\end{equation}
where $\lambda_{pr}$ and $\lambda_{cyc}$ are hyperparameters. 
We refer readers to the pseudocode and illustration for training in the supplementary material for more details.

\section{Experiments}

\subsection{Experiment Setup}


We utilize MaTLD (pretrained with LaMP \citep{li2024lamp}) as the generation network and train the style network from scratch. The style motion encoder in Figure~\ref{fig:contrastive} consists of a single transformer encoder, and the adaptor is a simple MLP. During the training of the style control network, the style network composed of four Transformer Encoder blocks is optimized while the parameters of the MaTLD are frozen. The framework is trained with the AdamW optimizer~\cite{loshchilov2017decoupled} and the learning rate is 1e-5. 
We evaluate motion generation, stylized motion generation, and motion style transfer on HumanML3D~\cite{Guo_2022_CVPR}, KIT-ML~\cite{KIT}, 100STYLE~\cite{mason2022local}, and Xia~\citep{xia2015realtime} datasets.
More details about the datasets, metrics, and baselines can be found in the supplementary.

\begin{table}[]
\centering
\small
\setlength{\tabcolsep}{1.5mm}{
\begin{tabular}{lcccc}
\toprule
\multirow{2}{*}{Method}&\multirow{2}{*}{SRA(\%)$\uparrow$}&\multirow{2}{*}{FID$\downarrow$}&Foot Skate\\
&&&Ratio $\downarrow$\\
\midrule
MLD+Aberman \citep{aberman2020unpaired}&61.006&3.892&0.338\\
MLD+Motion Puzzle \citep{jang2022motion}&67.233&6.871&0.197\\
SMooDi \citep{zhong2025smoodi}&65.147&1.582&0.095\\
BiFlow + MLD&\cellcolor{color1}70.238&\cellcolor{color1}1.566&\cellcolor{color2}0.087\\
BiFlow + MaTLD (Ours)&\cellcolor{color2}71.327&\cellcolor{color2}{1.497}&\cellcolor{color1}{0.093}\\
\bottomrule
\end{tabular}}
\vspace{-2mm}
\caption{{Quantitative results of motion style transfer on the HumanML3D dataset.}}
\vspace{-4mm}
\label{humanml}
\end{table}

\begin{table}[]
\centering
\small
\setlength{\tabcolsep}{1mm}{
\begin{tabular}{lcccc}
\toprule
\multirow{2}{*}{Method} & \multirow{2}{*}{\shortstack{SRA\\(\%)$\uparrow$}} & \multirow{2}{*}{\shortstack{CRA\\(\%)$\uparrow$}} & \multirow{2}{*}{FID$\downarrow$} & Foot Skate\\
&&&&Ratio $\downarrow$\\
\midrule
MLD+Aberman \citep{aberman2020unpaired}&56.667&34.444&5.681&\cellcolor{color2}{0.0260}\\
MLD+Motion Puzzle \citep{jang2022motion}&67.778&25.556&5.360&0.0316\\
SMooDi \citep{zhong2025smoodi}& 61.111&45.555&4.663&0.0317\\
BiFlow + MLD&\cellcolor{color1}67.870&\cellcolor{color1}{47.071}&\cellcolor{color1}{4.582}&\cellcolor{color1}0.0304\\
BiFlow + MaTLD (Ours)&\cellcolor{color2}67.962&\cellcolor{color2}47.635&\cellcolor{color2}4.533&\cellcolor{color1}0.0304\\
\bottomrule
\end{tabular}}
\vspace{-2mm}
\caption{Quantitative results of motion style transfer on the Xia dataset. We report the Content Recognition Accuracy (CRA).}
\vspace{-4mm}
\label{xia}
\end{table}

\begin{table*}[t]
\small
\centering
{
\begin{tabular}{ccccccccc}
    \toprule
    \multirow{2}{*}{} & \multirow{2}{*}{Methods} &\multicolumn{3}{c}{R Precision$\uparrow$} & \multirow{2}{*}{FID$\downarrow$} & \multirow{2}{1.5cm}{\centering MultiModal \\ Dist$\downarrow$} & \multirow{2}{1.5cm}{\centering Diversity $\rightarrow$}\\
    \cline{3-5} 
    &&{Top 1} & {Top 2} & {Top 3}  \\
    \midrule
    Dataset&Ground Truth &$0.511^{\pm.003}$ & $0.703^{\pm.003}$ & $0.797^{\pm.002}$ & $0.002^{\pm.00}$ & $2.974^{\pm.008}$ &$9.503^{\pm.065}$\\
    \midrule
    \multirow{2}{*}{{HumanML3D}}&MLD \citep{chen2023executing}&\cellcolor{color1}$0.481^{\pm.003}$ & \cellcolor{color1}$0.673^{\pm.003}$ & \cellcolor{color1}$0.772^{\pm.002}$ & \cellcolor{color1}$0.473^{\pm.013}$ &\cellcolor{color1}$3.196^{\pm.010}$ & \cellcolor{color1}\cellcolor{color1}$9.724^{\pm.082}$\\
    &{MaTLD \textit{w. LaMP}}&\cellcolor{color2}{$0.542^{\pm.003}$} & \cellcolor{color2}$0.733^{\pm.002}$ & \cellcolor{color2}{$0.828^{\pm.001}$ }& \cellcolor{color2}$0.152^{\pm.010}$ & \cellcolor{color2}$2.878^{\pm.006}$ & \cellcolor{color2}$9.703^{\pm.078}$\\    
    \midrule
    \midrule
    Dataset&Ground Truth &$0.424^{\pm.005}$ & $0.649^{\pm.006}$ & $0.779^{\pm.006}$ & $0.031^{\pm.004}$ & $2.788^{\pm.012}$ &$11.080^{\pm.097}$\\
        \midrule
    \multirow{2}{*}{{KIT-ML}}&MLD \citep{chen2023executing}&\cellcolor{color1}$0.390^{\pm.008}$ & \cellcolor{color1}$0.609^{\pm.008}$ &\cellcolor{color1} $0.734^{\pm.007}$ & \cellcolor{color1}$0.404^{\pm.027}$ & \cellcolor{color1}$3.204^{\pm.027}$ &\cellcolor{color2} $10.80^{\pm.117}$\\
    &{MaTLD \textit{w. LaMP}}&\cellcolor{color2}{$0.454^{\pm.007}$} & \cellcolor{color2}{$0.653^{\pm.006}$} &\cellcolor{color2} $0.771^{\pm.006}$ & \cellcolor{color2}$0.294^{\pm.015}$ & \cellcolor{color2}$2.909^{\pm.024}$ &\cellcolor{color1} $10.623^{\pm.107}$\\
    \bottomrule
\end{tabular}
}
\vspace{-3mm}
\caption{The quantitative results of our MaTLD with evaluator following previous methods on the HumanML3D dataset and the KIT-ML dataset, compared with MLD.}
\vspace{-4mm}
\label{tab:t2m}
\end{table*}

\setlength\tabcolsep{2pt}
\begin{table}[]
\centering
\small
\begin{tabular}{lccccc}
\toprule
\multirow{2}{*}{Framework}&\multirow{2}{*}{SRA(\%)$\uparrow$}&\multirow{2}{*}{FID$\downarrow$}&\multirow{2}{*}{MM Dist$\downarrow$}& R-Prec $\uparrow$ &Foot skating\\
&&&&(Top 3)&ratio $\downarrow$\\
\midrule
(a) &72.418&\cellcolor{color1}1.609&\cellcolor{color1}4.477&\cellcolor{color1}0.571&\cellcolor{color1}0.124\\
(b) \color{color3}{\checkmark} &\cellcolor{color1}77.042&\cellcolor{color2}1.527&\cellcolor{color2}{4.292}&\cellcolor{color2}{0.613}&\cellcolor{color2}{0.118}\\
(c) &\cellcolor{color2}78.623&3.620&5.766&0.393&0.142\\
(d) &51.527&4.869&5.118&0.508&0.129\\
\bottomrule
\end{tabular}
\vspace{-2mm}
\caption{Ablation study on different control flow. For (b), (c), and (d), we utilize the same guidance scale and style guidance scale. In contrast, (a) employs the optimal guidance scale and style guidance scale as described in the SMooDi.}
\vspace{-3mm}
\label{tab:ablation}
\end{table}
\setlength\tabcolsep{6pt}


\subsection{Evaluation}

\noindent \textbf{Motion-guided stylized motion generation.}
For the stylized text-to-motion generation task with motions as style, we present a comparative analysis of our approach against four baseline methods, as summarized in Table \ref{stylegen}. When we input style motion as the control signal, our model outperforms existing methods on both SRA and FID metrics. SRA and FID are two trade-off metrics; however, compared to SMooDi~\citep{zhong2025smoodi}, our approach achieves an FID that is lower by 6.1\% while maintaining an SRA that is 7.7\% higher. As illustrated in Figure \ref{fig:res_compare}, SMooDi exhibits conflicts between style and content, resulting in the stylized motion failing to align with the content prompts. Furthermore, the style motion tends to lose some dynamic qualities, leading to inaccuracies in classifying stylized actions for similar poses, such as `Flapping' and `Aeroplane'.

\noindent \textbf{Text/image-guided stylized motion generation.}
When the control signal comes to texts, ChatGPT+MLD utilizes ChatGPT to merge style labels from 100STYLE with texts from HumanML3D into a sentence. For example, `a person walks’ + `old’ = `an elderly person walks’. Then this merged sentence is fed to MLD. The performance of ChatGPT+MLD is limited, achieving merely 4.819\% in the SRA metric. This suggests that the MLD framework, despite incorporating style descriptors, falls short in facilitating stylized generation from text inputs alone. In contrast, our Adaptor+MLD/MaTLD accepts one label word as signal control, and fully leverages the control mechanism of the \methodName, improving the SRA performance from 4.819\% to 52.1\% and 52.307\% with only a small degradation on the FID.
The visualization of the generated stylized motions from texts or images is displayed in Figure~\ref{fig:res_multimodal}, from where we can see the convenient multimodal control.

\noindent \textbf{Motion style transfer.}
For the motion style transfer task, we follow the settings of SMooDi and evaluate only three metrics: FID, SRA, and Foot skating ratio. We utilize the HumanML3D dataset~\citep{Guo_2022_CVPR} as the source of motion content and extract motion styles from the 100STYLE dataset. As shown in Table \ref{humanml}, our algorithm achieves a significant improvement of 9.4\% on the SRA metric, along with advancements of 5.8\% and 8.4\% on the FID and Foot skating ratio metrics, respectively. Furthermore, we conduct experiments on the Xia dataset~\citep{xia2015realtime}, which was not seen during the training phase of both our model and the baseline model. Given the presence of motion content labels in the Xia dataset, we report the Content Recognition Accuracy (CRA). The results are presented in Table \ref{xia}.
 
\noindent \textbf{Text-to-Motion Generation.}
In Table \ref{tab:t2m}, we report the text-to-motion results using CLIP and LaMP as text encoders, comparing them with MLD. We employ our proposed motion-aligned temporal VAE encoding to obtain the motion latent representations, and the framework of the diffusion process is consistent with MLD. As shown in the table, there is a significant improvement compared to MLD, such as an improvement of 6.9\%, 5.3\%, and 4.7\% in R-Precision Top {1, 2, 3} and improves FID by 28.9\% on HumanML3D,
indicating that our enhancements to the VAE are both reasonable and effective.

\subsection{Ablation Study}
To validate the effectiveness of our framework, we conduct ablation studies on different control-feedback flows. As shown in Figure~\ref{fig:ablation}, we evaluate four architectures: (a), (b), (c), and (d). The ablation results are shown in Table~\ref{fig:ablation}.

In Figure~\ref{fig:ablation} (a), we align our approach with SMooDi, introducing a unidirectional control signal between the encoder of the generation network and the encoder of the style network, with results corresponding to the first row in Table~\ref{tab:ablation}. In Figure~\ref{fig:ablation} (b), we implement a bidirectional control flow between these two encoders, where the style signal influences the content, and at the same time, the content signal provides feedback to the style. We choose this as the final framework. In Figures~\ref{fig:ablation} (c) and \ref{fig:ablation} (d), bidirectional signals are also incorporated between the middle block and the decoder, respectively. These two frameworks lead to a degradation in the final results. We attribute this to the control signals from the style network disrupting the parameters of the frozen middle block and decoder of the generation network, thereby affecting the recovery of fine details.

\vspace{-1mm}

\section{Conclusion}

This paper proposes a motion-aligned temporal VAE and a multimodal stylized motion generation framework that innovatively leverages style motion, text, or image as style control to produce dynamic motion sequences. 
Different from previous methods where the control information only moves from style to content, a bidirectional control flow between the style network and the generation network is also introduced to resolve the conflict between the two spaces for better stylized motion generation.
\paragraph{Limitations}
Although the conflict between the content and style is relieved, the level of similarity between the generated motion and the style, and that between the generated motion and the content text is still a trade-off.

{
    \small
    \bibliographystyle{ieeenat_fullname}
    \bibliography{main}

\begin{thebibliography}{50}
\providecommand{\natexlab}[1]{#1}
\providecommand{\url}[1]{\texttt{#1}}
\expandafter\ifx\csname urlstyle\endcsname\relax
  \providecommand{\doi}[1]{doi: #1}\else
  \providecommand{\doi}{doi: \begingroup \urlstyle{rm}\Url}\fi

\bibitem[Aberman et~al.(2020)Aberman, Weng, Lischinski, Cohen-Or, and Chen]{aberman2020unpaired}
Kfir Aberman, Yijia Weng, Dani Lischinski, Daniel Cohen-Or, and Baoquan Chen.
\newblock Unpaired motion style transfer from video to animation.
\newblock \emph{{ACM Transactions on Graphics}}, 2020.

\bibitem[Andreou et~al.(2024)Andreou, Wang, Abrevaya, Cani, Chrysanthou, and Kalogeiton]{andreou2024lead}
Nefeli Andreou, Xi Wang, Victoria~Fern{\'a}ndez Abrevaya, Marie-Paule Cani, Yiorgos Chrysanthou, and Vicky Kalogeiton.
\newblock Lead: Latent realignment for human motion diffusion.
\newblock \emph{arXiv preprint arXiv:2410.14508}, 2024.

\bibitem[Ao et~al.(2023)Ao, Zhang, and Liu]{Ao2023GestureDiffuCLIP}
Tenglong Ao, Zeyi Zhang, and Libin Liu.
\newblock Gesturediffuclip: Gesture diffusion model with clip latents.
\newblock \emph{{ACM Transactions on Graphics}}, 2023.

\bibitem[Cen et~al.(2024)Cen, Pi, Peng, Shen, Yang, Zhu, Bao, and Zhou]{cen2024generating}
Zhi Cen, Huaijin Pi, Sida Peng, Zehong Shen, Minghui Yang, Shuai Zhu, Hujun Bao, and Xiaowei Zhou.
\newblock Generating human motion in 3d scenes from text descriptions.
\newblock In \emph{Proceedings of the IEEE/CVF Conference on Computer Vision and Pattern Recognition}, pages 1855--1866, 2024.

\bibitem[Chen et~al.(2024)Chen, Dai, Ju, Lu, and Zhang]{chen2024motionclr}
Ling-Hao Chen, Wenxun Dai, Xuan Ju, Shunlin Lu, and Lei Zhang.
\newblock Motionclr: Motion generation and training-free editing via understanding attention mechanisms.
\newblock \emph{arXiv preprint arXiv:2410.18977}, 2024.

\bibitem[Chen et~al.(2023)Chen, Jiang, Liu, Huang, Fu, Chen, and Yu]{chen2023executing}
Xin Chen, Biao Jiang, Wen Liu, Zilong Huang, Bin Fu, Tao Chen, and Gang Yu.
\newblock Executing your commands via motion diffusion in latent space.
\newblock In \emph{Proceedings of the IEEE Conference on Computer Vision and Pattern Recognition}, 2023.

\bibitem[Christen et~al.(2024)Christen, Hampali, Sener, Remelli, Hodan, Sauser, Ma, and Tekin]{christen2024diffh2o}
Sammy Christen, Shreyas Hampali, Fadime Sener, Edoardo Remelli, Tomas Hodan, Eric Sauser, Shugao Ma, and Bugra Tekin.
\newblock Diffh2o: Diffusion-based synthesis of hand-object interactions from textual descriptions.
\newblock In \emph{SIGGRAPH Asia 2024 Conference Papers}, 2024.

\bibitem[Dai et~al.(2024)Dai, Chen, Wang, Liu, Dai, and Tang]{dai2024motionlcm}
Wenxun Dai, Ling-Hao Chen, Jingbo Wang, Jinpeng Liu, Bo Dai, and Yansong Tang.
\newblock Motionlcm: Real-time controllable motion generation via latent consistency model.
\newblock \emph{arXiv preprint arXiv:2404.19759}, 2024.

\bibitem[Dhariwal and Nichol(2021)]{dhariwal2021diffusion}
Prafulla Dhariwal and Alexander Nichol.
\newblock Diffusion models beat gans on image synthesis.
\newblock \emph{Advances in Neural Information Processing Systems}, 2021.

\bibitem[Gatys et~al.(2016)Gatys, Ecker, and Bethge]{gatys2016image}
Leon~A Gatys, Alexander~S Ecker, and Matthias Bethge.
\newblock Image style transfer using convolutional neural networks.
\newblock In \emph{Proceedings of the IEEE Conference on Computer Vision and Pattern Recognition}, 2016.

\bibitem[Guo et~al.(2022)Guo, Zou, Zuo, Wang, Ji, Li, and Cheng]{Guo_2022_CVPR}
Chuan Guo, Shihao Zou, Xinxin Zuo, Sen Wang, Wei Ji, Xingyu Li, and Li Cheng.
\newblock Generating diverse and natural 3d human motions from text.
\newblock In \emph{Proceedings of the IEEE Conference on Computer Vision and Pattern Recognition}, 2022.

\bibitem[Guo et~al.(2024{\natexlab{a}})Guo, Mu, Javed, Wang, and Cheng]{guo2024momask}
Chuan Guo, Yuxuan Mu, Muhammad~Gohar Javed, Sen Wang, and Li Cheng.
\newblock Momask: Generative masked modeling of 3d human motions.
\newblock In \emph{Proceedings of the IEEE Conference on Computer Vision and Pattern Recognition}, 2024{\natexlab{a}}.

\bibitem[Guo et~al.(2024{\natexlab{b}})Guo, Mu, Zuo, Dai, Yan, Lu, and Cheng]{guo2024generative}
Chuan Guo, Yuxuan Mu, Xinxin Zuo, Peng Dai, Youliang Yan, Juwei Lu, and Li Cheng.
\newblock Generative human motion stylization in latent space.
\newblock \emph{arXiv preprint arXiv:2401.13505}, 2024{\natexlab{b}}.

\bibitem[He et~al.(2020)He, Fan, Wu, Xie, and Girshick]{he2020momentum}
Kaiming He, Haoqi Fan, Yuxin Wu, Saining Xie, and Ross Girshick.
\newblock Momentum contrast for unsupervised visual representation learning.
\newblock In \emph{Proceedings of the IEEE/CVF conference on computer vision and pattern recognition}, pages 9729--9738, 2020.

\bibitem[Hu et~al.(2021)Hu, Shen, Wallis, Allen-Zhu, Li, Wang, Wang, and Chen]{hu2021lora}
Edward~J Hu, Yelong Shen, Phillip Wallis, Zeyuan Allen-Zhu, Yuanzhi Li, Shean Wang, Lu Wang, and Weizhu Chen.
\newblock Lora: Low-rank adaptation of large language models.
\newblock \emph{arXiv preprint arXiv:2106.09685}, 2021.

\bibitem[Huang and Belongie(2017)]{huang2017arbitrary}
Xun Huang and Serge Belongie.
\newblock Arbitrary style transfer in real-time with adaptive instance normalization.
\newblock In \emph{Proceedings of the International Conference on Computer Vision}, 2017.

\bibitem[Jang et~al.(2022)Jang, Park, and Lee]{jang2022motion}
Deok-Kyeong Jang, Soomin Park, and Sung-Hee Lee.
\newblock Motion puzzle: Arbitrary motion style transfer by body part.
\newblock \emph{{ACM Transactions on Graphics}}, 2022.

\bibitem[Jiang et~al.(2023)Jiang, Chen, Liu, Yu, Yu, and Chen]{jiang2023motiongpt}
Biao Jiang, Xin Chen, Wen Liu, Jingyi Yu, Gang Yu, and Tao Chen.
\newblock Motiongpt: Human motion as a foreign language.
\newblock \emph{Advances in Neural Information Processing Systems}, 36:\penalty0 20067--20079, 2023.

\bibitem[Jones et~al.(2024)Jones, Wang, Kumari, Bau, and Zhu]{jones2024customizing}
Maxwell Jones, Sheng-Yu Wang, Nupur Kumari, David Bau, and Jun-Yan Zhu.
\newblock Customizing text-to-image models with a single image pair.
\newblock \emph{arXiv preprint arXiv:2405.01536}, 2024.

\bibitem[Karunratanakul et~al.(2023)Karunratanakul, Preechakul, Suwajanakorn, and Tang]{karunratanakul2023gmd}
Korrawe Karunratanakul, Konpat Preechakul, Supasorn Suwajanakorn, and Siyu Tang.
\newblock Gmd: Controllable human motion synthesis via guided diffusion models.
\newblock In \emph{Proceedings of the International Conference on Computer Vision}, 2023.

\bibitem[Kim et~al.(2024)Kim, Kim, Chang, and Choi]{kim2024most}
Boeun Kim, Jungho Kim, Hyung~Jin Chang, and Jin~Young Choi.
\newblock Most: Motion style transformer between diverse action contents.
\newblock In \emph{Proceedings of the IEEE/CVF Conference on Computer Vision and Pattern Recognition}, pages 1705--1714, 2024.

\bibitem[Kim and Lee(2019)]{kim2019perceptual}
Hye~Ji Kim and Sung-Hee Lee.
\newblock Perceptual characteristics by motion style category.
\newblock In \emph{Eurographics (Short Papers)}, 2019.

\bibitem[Kingma et~al.(2013)Kingma, Welling, et~al.]{kingma2013auto}
Diederik~P Kingma, Max Welling, et~al.
\newblock Auto-encoding variational bayes, 2013.

\bibitem[Li et~al.(2024)Li, Yuan, He, Qiu, Zhu, Gu, Shen, Dong, Dong, and Yang]{li2024lamp}
Zhe Li, Weihao Yuan, Yisheng He, Lingteng Qiu, Shenhao Zhu, Xiaodong Gu, Weichao Shen, Yuan Dong, Zilong Dong, and Laurence~T Yang.
\newblock Lamp: Language-motion pretraining for motion generation, retrieval, and captioning.
\newblock \emph{arXiv preprint arXiv:2410.07093}, 2024.

\bibitem[Loshchilov(2017)]{loshchilov2017decoupled}
I Loshchilov.
\newblock Decoupled weight decay regularization.
\newblock \emph{arXiv preprint arXiv:1711.05101}, 2017.

\bibitem[Mason et~al.(2022)Mason, Starke, and Komura]{mason2022local}
Ian Mason, Sebastian Starke, and Taku Komura.
\newblock Real-time style modelling of human locomotion via feature-wise transformations and local motion phases.
\newblock \emph{Proceedings of the ACM on Computer Graphics and Interactive Techniques}, 2022.

\bibitem[Park et~al.(2021)Park, Jang, and Lee]{park2021diverse}
Soomin Park, Deok-Kyeong Jang, and Sung-Hee Lee.
\newblock Diverse motion stylization for multiple style domains via spatial-temporal graph-based generative model.
\newblock \emph{Proceedings of the ACM on Computer Graphics and Interactive Techniques}, 2021.

\bibitem[Peng et~al.(2023)Peng, Xie, Wu, Jampani, Sun, and Jiang]{peng2023hoi}
Xiaogang Peng, Yiming Xie, Zizhao Wu, Varun Jampani, Deqing Sun, and Huaizu Jiang.
\newblock Hoi-diff: Text-driven synthesis of 3d human-object interactions using diffusion models.
\newblock \emph{arXiv preprint arXiv:2312.06553}, 2023.

\bibitem[Plappert et~al.(2016)Plappert, Mandery, and Asfour]{KIT}
Matthias Plappert, Christian Mandery, and Tamim Asfour.
\newblock The {KIT} motion-language dataset.
\newblock \emph{Big Data}, 4\penalty0 (4):\penalty0 236--252, 2016.

\bibitem[Qian et~al.(2024)Qian, Xiao, Wu, Yang, Li, Wang, Wang, Kou, and Zhang]{qian2024smcd}
Ziyun Qian, Zeyu Xiao, Zhenyi Wu, Dingkang Yang, Mingcheng Li, Shunli Wang, Shuaibing Wang, Dongliang Kou, and Lihua Zhang.
\newblock Smcd: High realism motion style transfer via mamba-based diffusion.
\newblock \emph{arXiv preprint arXiv:2405.02844}, 2024.

\bibitem[Raab et~al.(2023)Raab, Leibovitch, Tevet, Arar, Bermano, and Cohen-Or]{raab2023single}
Sigal Raab, Inbal Leibovitch, Guy Tevet, Moab Arar, Amit~H Bermano, and Daniel Cohen-Or.
\newblock Single motion diffusion.
\newblock \emph{arXiv preprint arXiv:2302.05905}, 2023.

\bibitem[Radford et~al.(2021)Radford, Kim, Hallacy, Ramesh, Goh, Agarwal, Sastry, Askell, Mishkin, Clark, et~al.]{radford2021learning}
Alec Radford, Jong~Wook Kim, Chris Hallacy, Aditya Ramesh, Gabriel Goh, Sandhini Agarwal, Girish Sastry, Amanda Askell, Pamela Mishkin, Jack Clark, et~al.
\newblock Learning transferable visual models from natural language supervision.
\newblock In \emph{Proceedings of the International Conference on Machine Learning}, 2021.

\bibitem[Rombach et~al.(2022)Rombach, Blattmann, Lorenz, Esser, and Ommer]{rombach2022high}
Robin Rombach, Andreas Blattmann, Dominik Lorenz, Patrick Esser, and Bj{\"o}rn Ommer.
\newblock High-resolution image synthesis with latent diffusion models.
\newblock In \emph{Proceedings of the IEEE Conference on Computer Vision and Pattern Recognition}, 2022.

\bibitem[Shah et~al.(2025)Shah, Ruiz, Cole, Lu, Lazebnik, Li, and Jampani]{shah2025ziplora}
Viraj Shah, Nataniel Ruiz, Forrester Cole, Erika Lu, Svetlana Lazebnik, Yuanzhen Li, and Varun Jampani.
\newblock Ziplora: Any subject in any style by effectively merging loras.
\newblock In \emph{Proceedings of the European Conference on Computer Vision}, pages 422--438. Springer, 2025.

\bibitem[Song et~al.(2020)Song, Meng, and Ermon]{song2020denoising}
Jiaming Song, Chenlin Meng, and Stefano Ermon.
\newblock Denoising diffusion implicit models.
\newblock \emph{arXiv preprint arXiv:2010.02502}, 2020.

\bibitem[Song et~al.(2024)Song, Jin, Li, Chen, Hao, Hou, Li, and Qin]{song2024arbitrary}
Wenfeng Song, Xingliang Jin, Shuai Li, Chenglizhao Chen, Aimin Hao, Xia Hou, Ning Li, and Hong Qin.
\newblock Arbitrary motion style transfer with multi-condition motion latent diffusion model.
\newblock In \emph{Proceedings of the IEEE/CVF Conference on Computer Vision and Pattern Recognition}, pages 821--830, 2024.

\bibitem[Tang et~al.(2023)Tang, Wu, Wang, Hu, Gong, Liao, Li, Kou, and Jin]{tang2023rsmt}
Xiangjun Tang, Linjun Wu, He Wang, Bo Hu, Xu Gong, Yuchen Liao, Songnan Li, Qilong Kou, and Xiaogang Jin.
\newblock Rsmt: Real-time stylized motion transition for characters.
\newblock In \emph{ACM SIGGRAPH 2023 Conference Proceedings}, pages 1--10, 2023.

\bibitem[Tao et~al.(2022)Tao, Zhan, Chen, and van~de Panne]{tao2022style}
Tianxin Tao, Xiaohang Zhan, Zhongquan Chen, and Michiel van~de Panne.
\newblock Style-erd: Responsive and coherent online motion style transfer.
\newblock \emph{Arxiv}, 2022.

\bibitem[Tevet et~al.(2023)Tevet, Raab, Gordon, Shafir, Cohen-or, and Bermano]{tevet2023human}
Guy Tevet, Sigal Raab, Brian Gordon, Yonatan Shafir, Daniel Cohen-or, and Amit~Haim Bermano.
\newblock Human motion diffusion model.
\newblock In \emph{Proceedings of the International Conference on Learning Representations}, 2023.

\bibitem[Wan et~al.(2023)Wan, Dou, Komura, Wang, Jayaraman, and Liu]{wan2023tlcontrol}
Weilin Wan, Zhiyang Dou, Taku Komura, Wenping Wang, Dinesh Jayaraman, and Lingjie Liu.
\newblock Tlcontrol: Trajectory and language control for human motion synthesis.
\newblock \emph{arXiv preprint arXiv:2311.17135}, 2023.

\bibitem[Wang et~al.(2024)Wang, Wang, Xie, Qi, Shan, Wang, and Luo]{wang2024styleadapter}
Zhouxia Wang, Xintao Wang, Liangbin Xie, Zhongang Qi, Ying Shan, Wenping Wang, and Ping Luo.
\newblock Styleadapter: A unified stylized image generation model.
\newblock \emph{International Journal of Computer Vision}, 2024.

\bibitem[Wen et~al.(2021)Wen, Yang, Fu, Gao, Sun, and Liu]{wen2021autoregressive}
Yu-Hui Wen, Zhipeng Yang, Hongbo Fu, Lin Gao, Yanan Sun, and Yong-Jin Liu.
\newblock Autoregressive stylized motion synthesis with generative flow.
\newblock In \emph{Proceedings of the IEEE Conference on Computer Vision and Pattern Recognition}, 2021.

\bibitem[Wu et~al.(2024)Wu, Shi, Huang, Yu, Xu, and Wang]{wu2024thor}
Qianyang Wu, Ye Shi, Xiaoshui Huang, Jingyi Yu, Lan Xu, and Jingya Wang.
\newblock Thor: Text to human-object interaction diffusion via relation intervention.
\newblock \emph{arXiv preprint arXiv:2403.11208}, 2024.

\bibitem[Xia et~al.(2015)Xia, Wang, Chai, and Hodgins]{xia2015realtime}
Shihong Xia, Congyi Wang, Jinxiang Chai, and Jessica Hodgins.
\newblock Realtime style transfer for unlabeled heterogeneous human motion.
\newblock \emph{{ACM Transactions on Graphics}}, 34\penalty0 (4):\penalty0 1--10, 2015.

\bibitem[Xie et~al.(2024)Xie, Jampani, Zhong, Sun, and Jiang]{xie2023omnicontrol}
Yiming Xie, Varun Jampani, Lei Zhong, Deqing Sun, and Huaizu Jiang.
\newblock Omnicontrol: Control any joint at any time for human motion generation.
\newblock In \emph{Proceedings of the International Conference on Learning Representations}, 2024.

\bibitem[Xu et~al.(2020)Xu, Xu, Ni, Yang, Wang, and Darrell]{xu2020hierarchical}
Jingwei Xu, Huazhe Xu, Bingbing Ni, Xiaokang Yang, Xiaolong Wang, and Trevor Darrell.
\newblock Hierarchical style-based networks for motion synthesis.
\newblock In \emph{Proceedings of the European Conference on Computer Vision}, 2020.

\bibitem[Yi et~al.(2025)Yi, Thies, Black, Peng, and Rempe]{yi2025generating}
Hongwei Yi, Justus Thies, Michael~J Black, Xue~Bin Peng, and Davis Rempe.
\newblock Generating human interaction motions in scenes with text control.
\newblock In \emph{European Conference on Computer Vision}, pages 246--263. Springer, 2025.

\bibitem[Yu et~al.(2023)Yu, Wang, Zhao, Ghanem, and Zhang]{yu2023freedom}
Jiwen Yu, Yinhuai Wang, Chen Zhao, Bernard Ghanem, and Jian Zhang.
\newblock Freedom: Training-free energy-guided conditional diffusion model.
\newblock In \emph{Proceedings of the IEEE/CVF International Conference on Computer Vision}, pages 23174--23184, 2023.

\bibitem[Zhang et~al.(2023)Zhang, Rao, and Agrawala]{zhang2023adding}
Lvmin Zhang, Anyi Rao, and Maneesh Agrawala.
\newblock Adding conditional control to text-to-image diffusion models.
\newblock In \emph{Proceedings of the International Conference on Computer Vision}, 2023.

\bibitem[Zhong et~al.(2024)Zhong, Xie, Jampani, Sun, and Jiang]{zhong2025smoodi}
Lei Zhong, Yiming Xie, Varun Jampani, Deqing Sun, and Huaizu Jiang.
\newblock Smoodi: Stylized motion diffusion model.
\newblock In \emph{Proceedings of the European Conference on Computer Vision}, 2024.

\end{thebibliography}
}


\clearpage
\appendix


\section{Introduction}
This is the supplementary material, which is divided into the following sections:

1. The details of dataset are shown in Section \ref{data}. 

2. The evaluation metrics are shown in Section \ref{metric}.

3. The baselines we use in the experiment are shown in Section \ref{baseline}.

4. We provide the pseudo code, which is presented in Algorithm \ref{alg:inference}.

5. The details of motion style transfer task are elaborated in Section \ref{mst}.

6. The inference time cost is provided in Section \ref{inference}.

7. We report a user study of our method in Section \ref{user study}.

8. More ablation studies are conducted in Section \ref{aas}.

9. More visualization results are presented in Section \ref{morevisual}, including additional qualitative results and some cases to validate the generalization ability.

\begin{table*}[]
\centering
\small
\begin{tabularx}{\textwidth}{|c|X|}
\hline
\textbf{Category} & \textbf{Label} \\ \hline
CHAR & Aeroplane, Cat, Chicken, Dinosaur, Fairy, Monk, Morris, Penguin, Quail, Roadrunner, Robot, Rocket, Star, Superman, Zombie (15) \\ \hline
PER & Balance, Heavyset, Old, Rushed, Stiff (5) \\ \hline
EMO & Angry, Depressed, Elated, Proud (4) \\ \hline
ACT & kimbo, ArmsAboveHead, ArmsBehindBack, ArmsBySide, ArmsFolded, BeatChest, BentForward, BentKnees, BigSteps, BouncyLeft, BouncyRight, CrossOver, FlickLegs, Followed, GracefulArms, HandsBetweenLegs, HandsInPockets, HighKnees, KarateChop, Kick, LeanBack, LeanLeft, LeanRight, LeftHop, LegsApart, LimpLeft, LimpRight, LookUp, Lunge, March, Punch, RaisedLeftArm, RaisedRightArm, RightHop, Skip, SlideFeet, SpinAntiClock, SpinClock, StartStop, Strutting, Sweep, Teapot, Tiptoe, TogetherStep, TwoFootJump, WalkingStickLeft, WalkingStickRight, Waving, WhirlArms, WideLegs, WiggleHips, WildArms, WildLegs (58) \\ \hline
MOT & CrowdAvoidance, InTheDark, LawnMower, OnHeels, OnPhoneLeft, OnPhoneRight, OnToesBentForward, OnToesCrouched, Rushed (9) \\ \hline
OBJ & DragLeftLeg, DragRightLeg, DuckFoot, Flapping, ShieldedLeft, ShieldedRight, Swimming, SwingArmsRound, SwingShoulders (9)\\ \hline
\end{tabularx}
\caption{The detailed grouping of style labels in the 100STYLE dataset.}
\label{tab:100style}
\end{table*}
\section{Datasets.}
\label{data}
We utilize the HumanML3D dataset~\cite{Guo_2022_CVPR} as the motion content dataset and the 100STYLE dataset~\cite{mason2022local} as the motion style dataset for training. 
Moreover, we use Xia dataset~\citep{xia2015realtime} as the style dataset for evaluating the motion style transfer task.
The HumanML3D dataset is the largest motion content dataset paired with text annotations and is widely used for text-to-motion generation.
The 100STYLE dataset~\cite{mason2022local} is the largest motion style dataset. Xia dataset is a smaller motion style collection encompassing 8 distinct styles with precise annotations for 8 action types. The motions within this collection are generally under 3 seconds in duration.
Due to differences in skeletons between the 100STYLE dataset and HumanML3D, we retarget the motions from 100STYLE to match the HumanML3D (SMPL) skeleton. Following this alignment, we apply the same processing steps as used for the HumanML3D dataset to preprocess the 100STYLE dataset.
We also follow SMooDi~\cite{zhong2025smoodi} and use MotionGPT~\cite{jiang2023motiongpt} to generate pseudo text descriptions for the motion sequences.
For both the HumanML3D dataset and the 100STYLE dataset, we use the consistent root-velocity motion representations outlined in~\cite{Guo_2022_CVPR}.


To avoid content conflicts between the content text and the style label, we filter certain style labels in the 100STYLE dataset during evaluation, following the approach of SMooDi~\cite{zhong2025smoodi}.
Specifically, following the approach in \cite{kim2019perceptual}, we categorize the style labels in the 100STYLE dataset into six groups: character (CHAR), personality (PER), emotion (EMO), action (ACT), objective (OBJ), and motivation (MOT). Notably, since the ACT group conveys content-related information, we exclude style motions from this group when calculating the SRA metric for content text derived from the HumanML3D dataset.
Table~\ref{tab:100style} shows the detailed grouping of style labels in the 100STYLE dataset.

\section{Evaluation Metrics.} 
\label{metric}
The metrics are designed to assess three key dimensions: Content Preservation, Style Reflection, and Realism.
For evaluating content preservation, we follow~\cite{chen2023executing} and use metrics including \textit{motion-retrieval precision (R-precision)}, \textit{Multi-modal Distance (MM Dist)}, \textit{Diversity}, and \textit{Frechet Inception Distance (FID)}.
For realism, we address common foot skating issues by integrating the \textit{Foot Skating Ratio} metric proposed by~\cite{karunratanakul2023gmd}.
For style reflection, we follow SMooDi~\cite{zhong2025smoodi} to use \textit{Style Recognition Accuracy (SRA)}~\cite{jang2022motion}. SRA and FID are two metrics that involve a trade-off. A higher SRA indicates more pronounced stylization but leads to a relatively worse FID. During the evaluation, a content text from the HumanML3D dataset and a motion style sequence from the 100STYLE dataset, are randomly selected to generate the stylized motion. We compute the SRA for the generated motion with a pretrainde style classifier.

\section{Baselines.}
\label{baseline}
We focus on three primary tasks: motion-guided stylized motion generation, motion style transfer, and text/image-guided stylized motion generation. For the motion-guided stylized motion generation and motion style transfer tasks, we utilize the evaluation metrics established in SMooDi~\cite{zhong2025smoodi}, incorporating MaTLD as the text-to-motion model; the methods for motion style transfer are consistent with those employed in the motion style transfer task. For the text/image-guided stylized motion generation task, we extract features from the texts and images using CLIP and input these features into our trained adaptor as style embeddings for style control. Unlike ChatGPT+MLD, which essentially functions as a straightforward text-to-motion model without embodying the control aspect, our approach allows for style control using a simple label word, enabling more nuanced manipulation of the generated motion styles.

\section{More Implementation Details}
\label{imple}
Our proposed framework is implemented on PyTorch and executed on eight NVIDIA V100 GPUs. The training process employs a batch size of 256 and spans 600 epochs, utilizing the AdamW optimizer~\cite{loshchilov2017decoupled} with a learning rate of 1e-5.  We utilize MaTLD as the pretrained generation network and train the style network from scratch. The style motion encoder in Figure 4 in the main paper
consists of a single transformer encoder, and the adaptor is a simple MLP trained for 500 epochs by contrastive learning.
During training, the style network composed of four Transformer Encoder blocks is optimized while the parameters of the MaTLD are frozen. We follow SMooDi to train both the unconditioned and conditioned models simultaneously. We randomly assign the content text $\vc=\phi$ and mask out 10\% of the style motion sequence $\mathbf{s}$ in the temporal dimension. The number of diffusion steps is 1000 during training and reduced to 50 during inference. The weight for classifier-free content guidance $w_c$ is set to 15, and the weight for classifier-free style guidance $w_s$ is set to 1.6. $w_s$ is the primary parameter that we adjust for balancing the style and content, and its impact is shown in the Figure \ref{fig:zhexian}. 


\begin{figure}[]
\centering
  \includegraphics[width=0.9\columnwidth, trim={0cm 0cm 0cm 0cm}, clip]{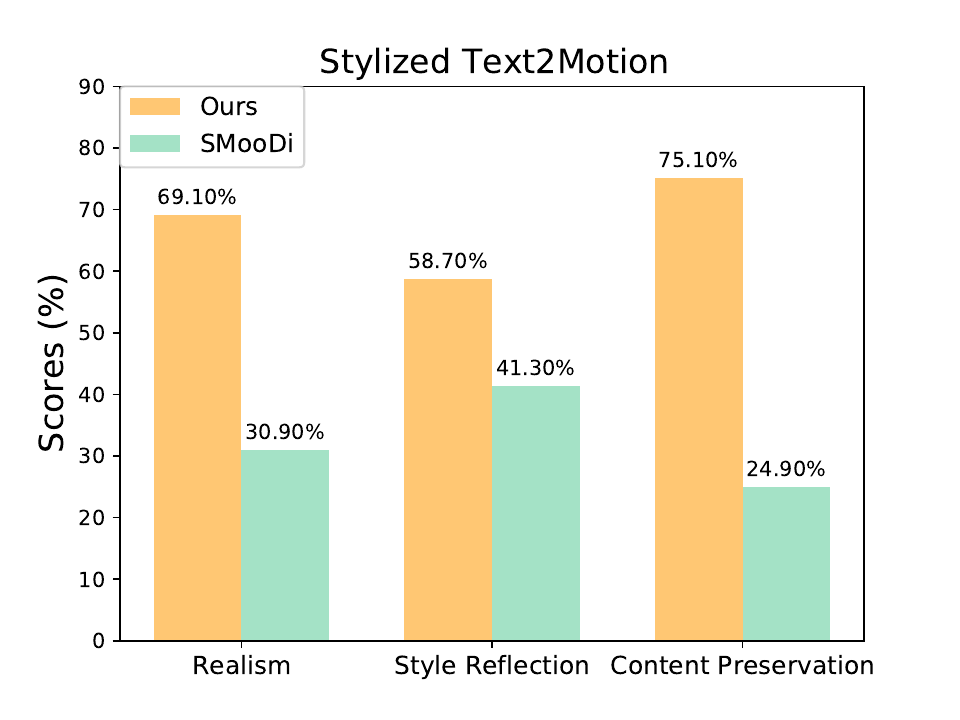}
\vspace{-5mm}
\caption{User study on the stylized motion generation task.
}
\label{fig:user_study}
\vspace{-3mm}
\end{figure}

\begin{table*}[]
\centering
\small
\setlength{\tabcolsep}{2.5mm}{
\begin{tabular}{lccc|ccccc}
\toprule
{Sub-}&\multirow{2}{*}{MaTLD}& MaTLD+ & MaTLD+ &Full&\multirow{2}{*}{Ours}&\multirow{2}{*}{SMooDi}&MLD+&MLD+\\
Modules&&BiFlow&classifier-based&Model&&&Motion Puzzle&Aberman\\
\midrule
Time (s) &0.2157&2.4942& 0.5583& Time (s)& 3.041&3.1133&0.2420&0.2275\\
\bottomrule
\end{tabular}}
\caption{{Inference time. The Average Inference Time per Sentence (AITS) in seconds for baselines and each of our submodules is reported.}}
\label{tab:time}
\end{table*}

\begin{table*}[]
\centering
\scriptsize
\begin{tabular}{cclccccccc}
\toprule
    \multirow{2}{*}{Task} & \multirow{2}{*}{Dataset} &\multicolumn{1}{c}{\multirow{2}{*}{Method}} & \multirow{2}{*}{SRA(\%)$\uparrow$} & \multirow{2}{*}{CRA(\%)$\uparrow$}&\multirow{2}{*}{FID$\downarrow$}&\multirow{2}{*}{MM Dist$\downarrow$}& R-Precision $\uparrow$ &Foot Skate\\
&&&&&&&(Top 3)&Ratio $\downarrow$\\
\midrule
\multirow{3}{*}{\shortstack{Stylized\\Motion Generation}}&\multirow{3}{*}{HumanML3D}
&BiFlow + MLD&77.042&-&1.527&4.292&0.613&0.118\\
&&BiFlow + MaTLD \textit{w. CLIP} &\cellcolor{color1}77.442&-&\cellcolor{color1}1.521&\cellcolor{color1}4.074&\cellcolor{color1}0.629&\cellcolor{color1}0.117\\
&&BiFlow + MaTLD \textit{w. LaMP} &\cellcolor{color2}78.029&-&\cellcolor{color2}1.511&\cellcolor{color2}3.823&\cellcolor{color2}0.676&\cellcolor{color2}0.116\\
\midrule
\multirow{6}{*}{\shortstack{Style\\Transfer}}&\multirow{3}{*}{HumanML3D}
&BiFlow + MLD &70.238&-&1.566&-&-&\cellcolor{color2}0.087\\
&&BiFlow + MaTLD \textit{w. CLIP} &\cellcolor{color1}70.586&-&\cellcolor{color1}1.553&-&-&0.095\\
&&BiFlow + MaTLD \textit{w. LaMP} &\cellcolor{color2}71.327&-&\cellcolor{color2}1.497&-&-&\cellcolor{color1}0.093\\
\cline{2-9}
\noalign{\vspace{1pt}}
&\multirow{3}{*}{Xia}
&BiFlow + MLD &67.870&47.071&4.582&-&-&0.0304\\
&&BiFlow + MaTLD \textit{w. CLIP} &\cellcolor{color1}67.917&\cellcolor{color1}47.226&\cellcolor{color1}4.551&-&-&0.0304\\
&&BiFlow + MaTLD \textit{w. LaMP} &\cellcolor{color2}67.962&\cellcolor{color2}47.635&\cellcolor{color2}4.533&-&-&0.0304\\
\midrule
\multirow{6}{*}{\shortstack{Motion\\Generation}}&\multirow{3}{*}{HumanML3D}
&MLD&-&-&$0.473^{\pm.013}$&$3.196^{\pm.010}$&{$0.772^{\pm.002}$ }&-\\
&&{MaTLD \textit{w. CLIP}}&-&-&\cellcolor{color1} $0.323^{\pm.011}$&\cellcolor{color1}$3.037^{\pm.010}$&\cellcolor{color1}{$0.791^{\pm.001}$ }&-\\
&&{MaTLD \textit{w. LaMP}}&-&-&\cellcolor{color2}$0.152^{\pm.010}$&\cellcolor{color2}$2.878^{\pm.006}$& \cellcolor{color2}{$0.828^{\pm.001}$ }&-\\
\cline{2-9}
\noalign{\vspace{1pt}}
&\multirow{3}{*}{KIT-ML}
&MLD&-&-&$0.404^{\pm.027}$ &$3.204^{\pm.027}$& $0.737^{\pm.007}$&-\\
&&{MaTLD \textit{w. CLIP}}&-&-&\cellcolor{color1} $0.383^{\pm.020}$ & \cellcolor{color1}$3.187^{\pm.018}$&\cellcolor{color1} $0.747^{\pm.005}$&-\\
&&{MaTLD \textit{w. LaMP}}&-&-& \cellcolor{color2}$0.294^{\pm.015}$ &\cellcolor{color2}$2.909^{\pm.024}$ &\cellcolor{color2} $0.771^{\pm.006}$&-\\
\bottomrule
\end{tabular}
\vspace{-2mm}
\caption{Quantitative comparison between MLD and MaTLD. It can be observed that, despite employing CLIP as the text encoder for training text-to-motion, MaTLD consistently outperforms in various tasks due to MLD, thereby validating the efficacy of the motion-aligned temporal VAE.
}
\vspace{-3mm}
\label{tab:mavae}
\end{table*}
\section{Motion Style Transfer}
\label{mst}
This task aims to generate a stylized motion sequence by combining a content motion sequence and a style motion sequence. As with SMooDi~\cite{zhong2025smoodi}, MulSMo can support this task effortlessly without requiring additional training.
Firstly, the deterministic DDIM reverse process~\cite{song2020denoising} is applied to derive the noised latent code $\vz_T^{Inv}$ corresponding to the content motion sequence.
The reverse process can be represented at step $t$ as:
\begin{align}
\vz_{t+1}^{Inv} = & \sqrt{\frac{\alpha_{t+1}}{\alpha_t}} \Bigg( \vz_t^{Inv} + 
\Bigg( \sqrt{\frac{1}{\alpha_{t+1}}} - 1 \Bigg) \nonumber \\ 
& - \Bigg( \sqrt{\frac{1}{\alpha_t}} - 1 \Bigg) \Bigg) \cdot 
\varepsilon_{\theta}(\vz_t^{Inv}; t, \vc, \emptyset),
\end{align}
where $\alpha$ represents the noise scale.
$\vz_T^{Inv}$ can be obtained at the last reverse step $T$. 
Given the reduced number of denoising steps compared to the stylized text-to-motion process, we slightly increase the weights for style guidance. Specifically, the denoising process consists of 30 steps, with $w_s = 6.5$ and $\tau = -0.4$.
\noindent\newcommand{\commentt}[1]{\textcolor{black}{#1}}
\begin{minipage}{1.0\linewidth} 

\begin{algorithm}[H]
\caption{\textbf{MulSMo}'s inference}\label{alg:inference}
\begin{algorithmic}[1]
\Require A motion diffusion model $G$ with parameters $\theta_G$, a style adaptor model $S$ with parameters $\theta_S$, zero linear adaptor $L$ with parameters $\theta{_{L_{c2s}}}$ and $\theta{_{L_{s2c}}}$, style motion sequence $\vs$ (if any), content texts $\vc$ (if any), the number of encoder blocks $B$.
\State $\vz_T \sim \mathcal{N}(\mathbf{0}, \mathbf{I})$ \commentt{\small \# Sample from pure Gaussian distribution}
\ForAll{$t$ from $T$ to $1$}    
\ForAll{$b$ from $0$ to $B-1$}    
        \State $\vr_b \leftarrow S(\vz_t, t, \vc, \vs; \theta_A)$ \commentt{\small \indent \# Style control}
    \State $\vg_b \leftarrow G(\vx_t, t, \vc, \theta_M)$ \commentt{\small \indent \# Generation Feedback}
    \State $\vx_t \leftarrow \vx_t + L(\vr_b, \theta_{L_{s2c}}) $ \commentt{\small \indent \# \textbf{Control}} 
    \State $\vz_t \leftarrow \vz_t + L(\vg_b, \theta_{L_{c2s}})$ \commentt{\small \indent \# \textbf{Feedback}}
    \EndFor
\State $\epsilon_t \leftarrow G(\vx_t, t, \vc, \theta_M)$ \commentt{\small \indent \# Generation Network}
    \ForAll{$k$ from $1$ to $K$} \commentt{\small \indent \# \textbf{Classifier-based style guidance}}
        \State $\epsilon_t = \epsilon_t  + \tau \nabla_{\vz_t}G(\vz_t, t, \mathbf{s}) $
    \EndFor
    \State $\vz_{t-1} \sim \mathcal{S}\left(\vz_t, \epsilon_t,t \right)$ \commentt{\small \# $S(\cdot,\cdot,\cdot)$ represents the DDIM sampling method~\cite{dhariwal2021diffusion}.}
\EndFor

\State $\vx_0$ = $\mathbf{D}(\vz_0)$
\\
\Return $\vx_0$
\end{algorithmic}
\end{algorithm}

\end{minipage}

\section{Inference Time and Pseudo Code}
\label{inference}
We present the Average Inference Time per Sentence (AITS) measured in seconds to assess the inference efficiency of bidirectional control flow and our full model compared with baseline methods for stylized motion generation task, as detailed in Table~\ref{tab:time}. Following SMooDi~\citep{zhong2025smoodi}, the AITS calculation is performed with a batch size of 1, excluding the time required for model and dataset loading, utilizing an NVIDIA A5000 GPU. The left part of the table ablates the inference time of each sub-module in our model while the right part indicates the running time of the full model, including MaTLD, style adaptor, and classifier guidance. Additional, we provide the pseudo code of our framework in Algorithm~\ref{alg:inference} to better illustrate how we implement the proposed bidirectional control flow mechanism.

\section{User Study}
\label{user study}
In order to address the inherently subjective nature of stylized motion, we conduct user studies employing pairwise comparisons to systematically evaluate our method in the task of stylized motion generation. A total of 41 human subjects encompassing a wide range of academic backgrounds were recruited to participate in this study. During each trial, each participant will be provided with 24 video segments, each approximately 9 seconds in length, generated by our method and SMooDi, respectively. They are tasked with selecting their preferred video, taking into account three key dimensions: Realism, Style Reflection, and Content Preservation. As illustrated in Figure~\ref{fig:user_study}, our method garnered significantly higher user preference across all three dimensions when compared to the SMooDi.

\section{Additional Ablation Studies}
\label{aas}
\paragraph{Different text encoders in MaTLD}
To demonstrate that the performance gains of our method in stylized motion generation, style transfer, and motion generation tasks are not solely attributable to LaMP \citep{li2024lamp}, but also rely on the motion-aligned temporal VAE, we report in Table \ref{tab:mavae} the capabilities of MaTLD trained with CLIP across these three tasks. As evidenced by the results, MaTLD \textit{w. CLIP} consistently outperforms MLD in all tasks, thereby validating the effectiveness of our motion-aligned temporal VAE architecture.

\begin{figure*}[]
    \centering
    \begin{subfigure}[b]{0.33\textwidth}
        \centering
        \includegraphics[width=\textwidth]{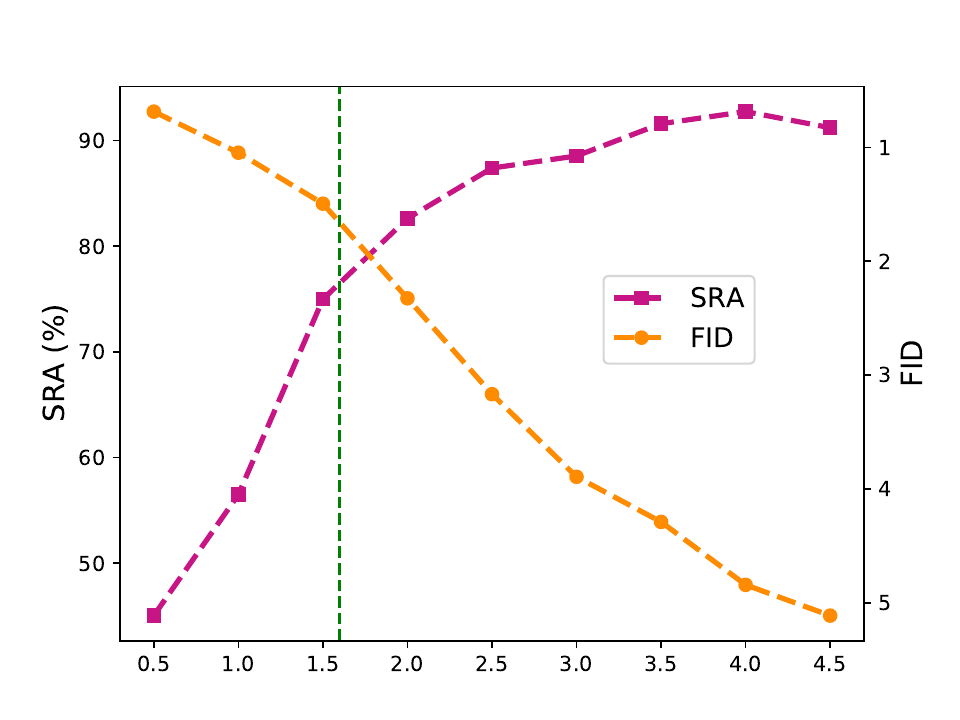} 
        \caption{Weights of $w_s$}
        \label{fig:sra_fid}
    \end{subfigure}
    \hfill
    \begin{subfigure}[b]{0.33\textwidth}
        \centering
        \includegraphics[width=\textwidth]{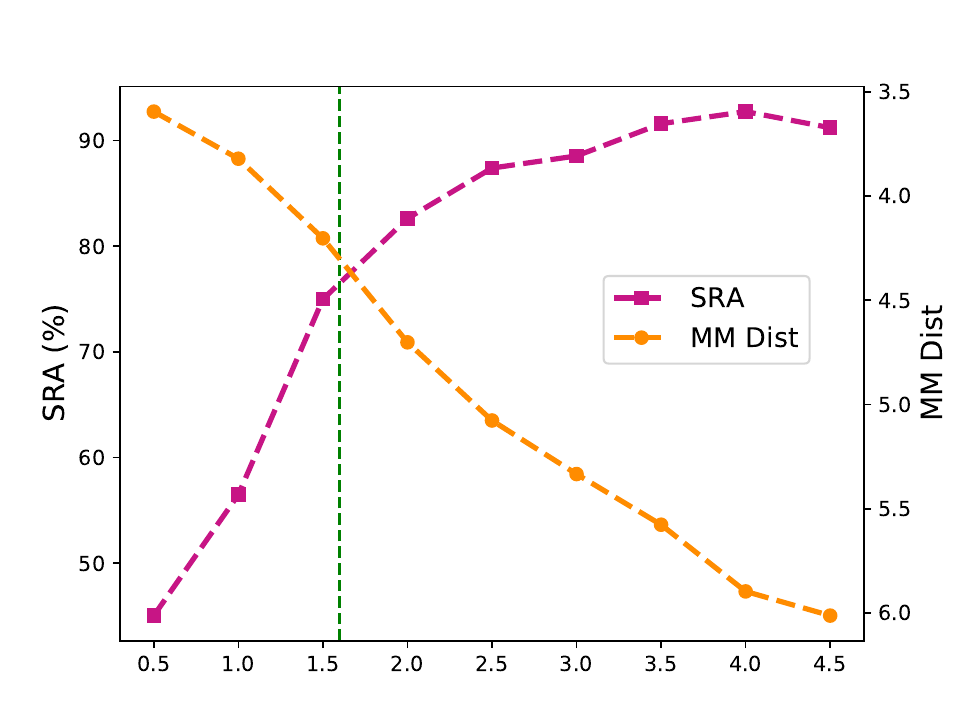}
        \caption{Weights of $w_s$}
        \label{fig:sra_mm}
    \end{subfigure}
    \hfill
    \begin{subfigure}[b]{0.33\textwidth}
        \centering
        \includegraphics[width=\textwidth]{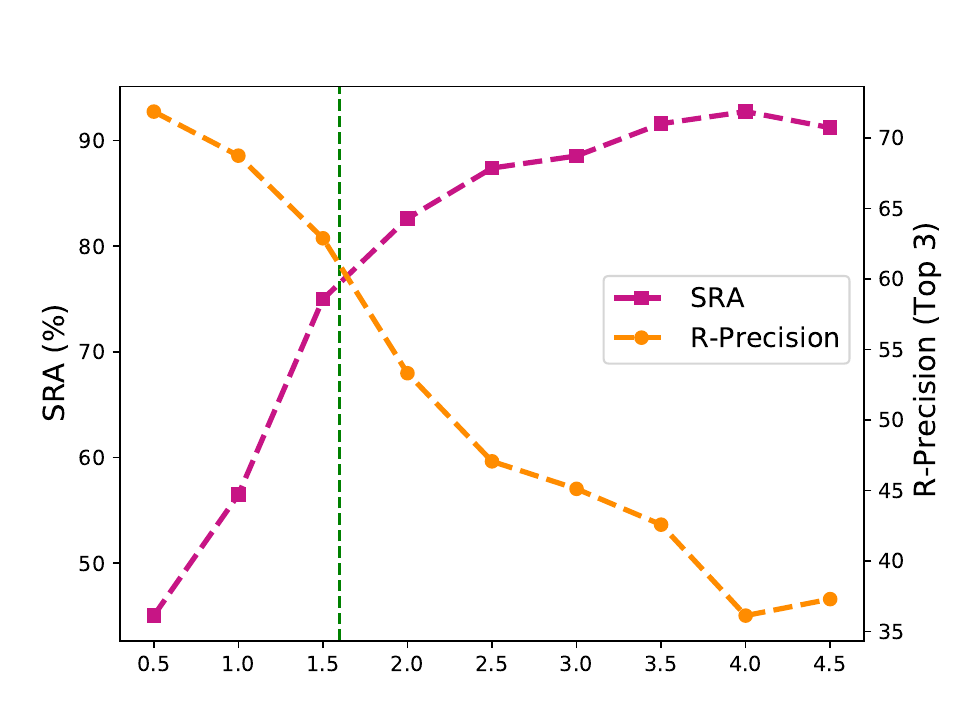} 
        \caption{Weights of $w_s$}
        \label{fig:sra_rp}
    \end{subfigure}
    \caption{Varying SRA, FID, R-Precision, and MM Dist under different values of classifier-free style guidance $w_s$.}
    \label{fig:zhexian}
\end{figure*}

\paragraph{Different weights of classifier-free style guidance.}
We find the weight of classifier-free style guidance $w_s$ has a huge impact on the results. In order to investigate the impact of $w_s$, we systematically vary these weights and present our findings in Figure~\ref{fig:zhexian}. Our results indicate that increasing the classifier-based style guidance weight enhances the SRA metric; however, this comes at the cost of reduced R Precision, MM Dist, and FID. These observations suggest that while the style is represented more accurately, there is notable content degradation. Finally, we set $w_s$ as 1.6. In this experiment, the weights of classifier-free content guidance $w_c$ and the weight of classifier-based style guidance $\tau$ are held constant, with $w_c=15, \tau=-0.2$.

\begin{figure*}[]
    \centering
    \begin{subfigure}[b]{0.33\textwidth}
        \centering
        \includegraphics[width=\textwidth]{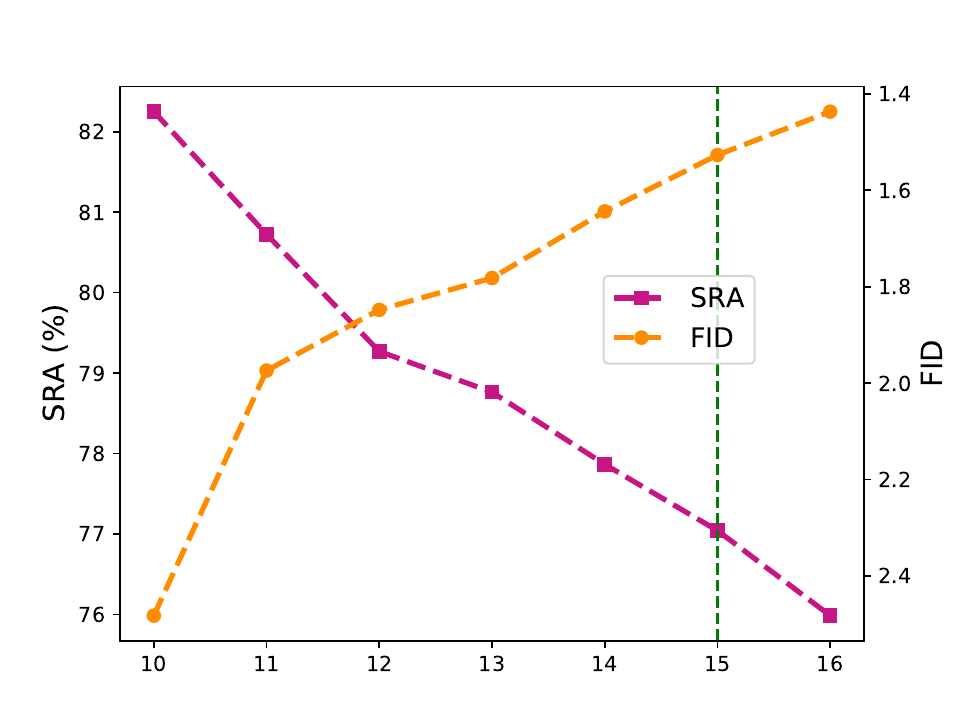} 
        \caption{Weights of $w_c$}
        \label{fig:sra_fid}
    \end{subfigure}
    \hfill
    \begin{subfigure}[b]{0.33\textwidth}
        \centering
        \includegraphics[width=\textwidth]{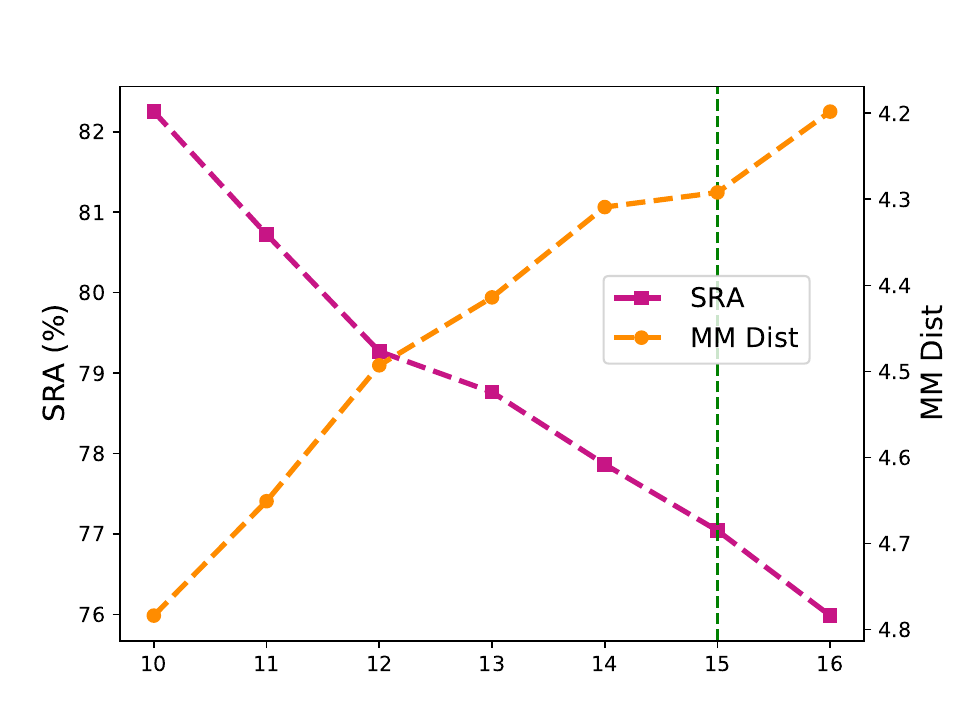}
        \caption{Weights of $w_c$}
        \label{fig:sra_mm}
    \end{subfigure}
    \hfill
    \begin{subfigure}[b]{0.33\textwidth}
        \centering
        \includegraphics[width=\textwidth]{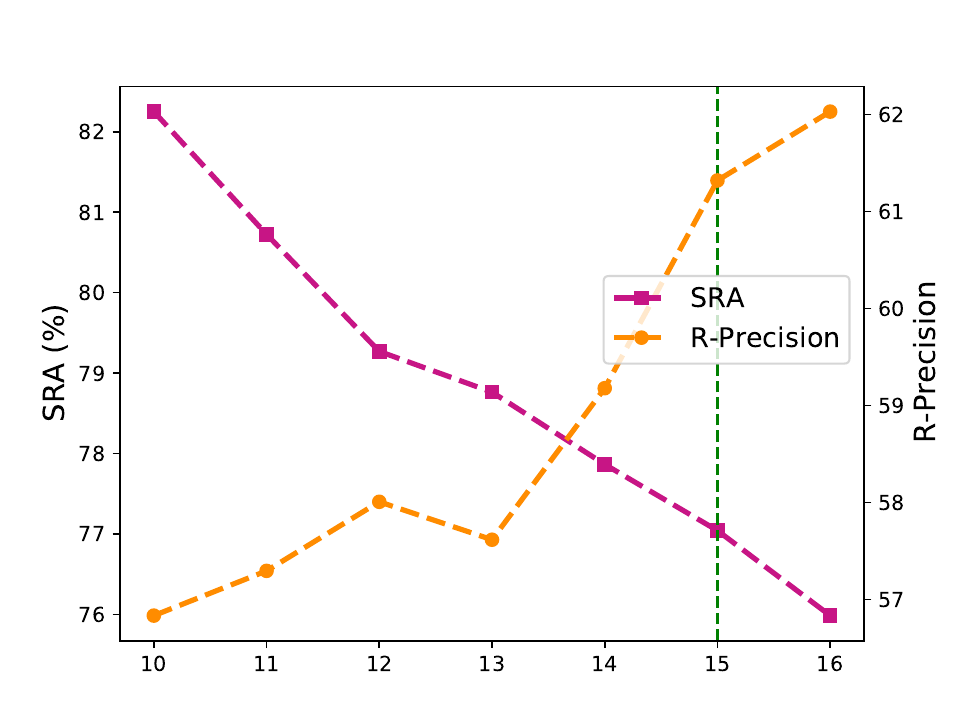} 
        \caption{Weights of $w_c$}
        \label{fig:sra_rp}
    \end{subfigure}
    \caption{Varying SRA, FID, R-Precision, and MM Dist under different values of classifier-free content guidance $w_c$.}
    \label{fig:wc}
\end{figure*}

\paragraph{Different weights of classifier-free content guidance.}
We also observe a significant impact of the weight of classifier-free content guidance $w_c$ on the results. As the value of $w_c$ increases, the model's content preservation capability improves; however, the style accuracy tends to deteriorate gradually. We present these weights and our findings in Figure~\ref{fig:wc}. Finally, we set $w_c$ as 15. In this experiment, the weights of classifier-free style guidance $w_s$ and the weight of classifier-based style guidance $\tau$ are held constant, with $w_s=1.6, \tau=-0.2$.

\section{More Visualization Results}
\label{morevisual}
\begin{figure*}
	\centering 
	\includegraphics[scale=0.6]{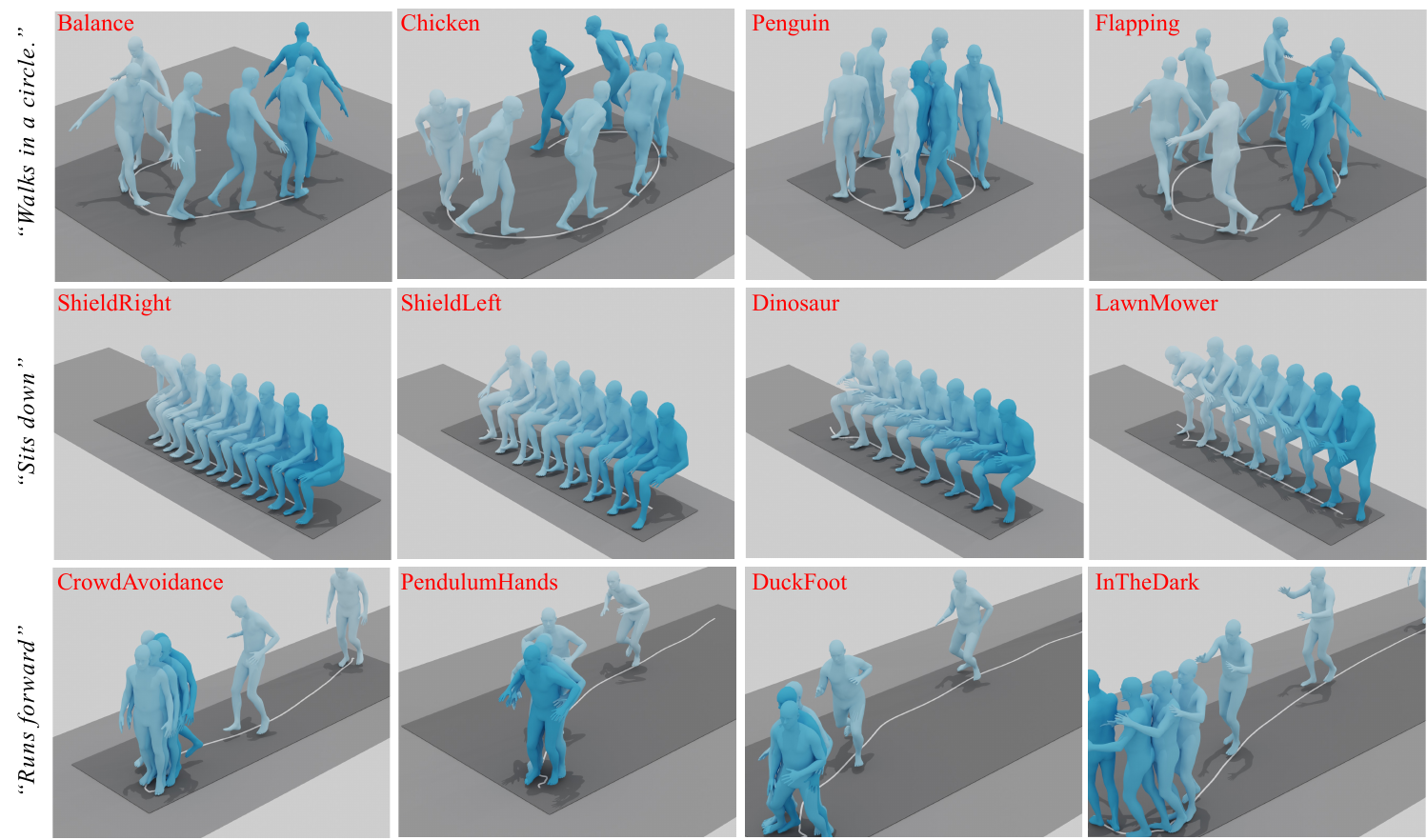} 
	\caption{More qualitative results of stylized motion generation on the 100STYLE dataset. The style name is annotated on the top-left of each image in red. The leftmost part of each row is the content prompt for that row.}  
	\label{fig:more_results} 
\end{figure*}

We present additional results of stylized motion generation as illustrated in Figure \ref{fig:more_results}. We select three content prompts, and for each content, we employ four distinct style motions as control signals, showcasing a total of 12 styles. 

\begin{figure*}
	\centering 
	\includegraphics[scale=0.5]{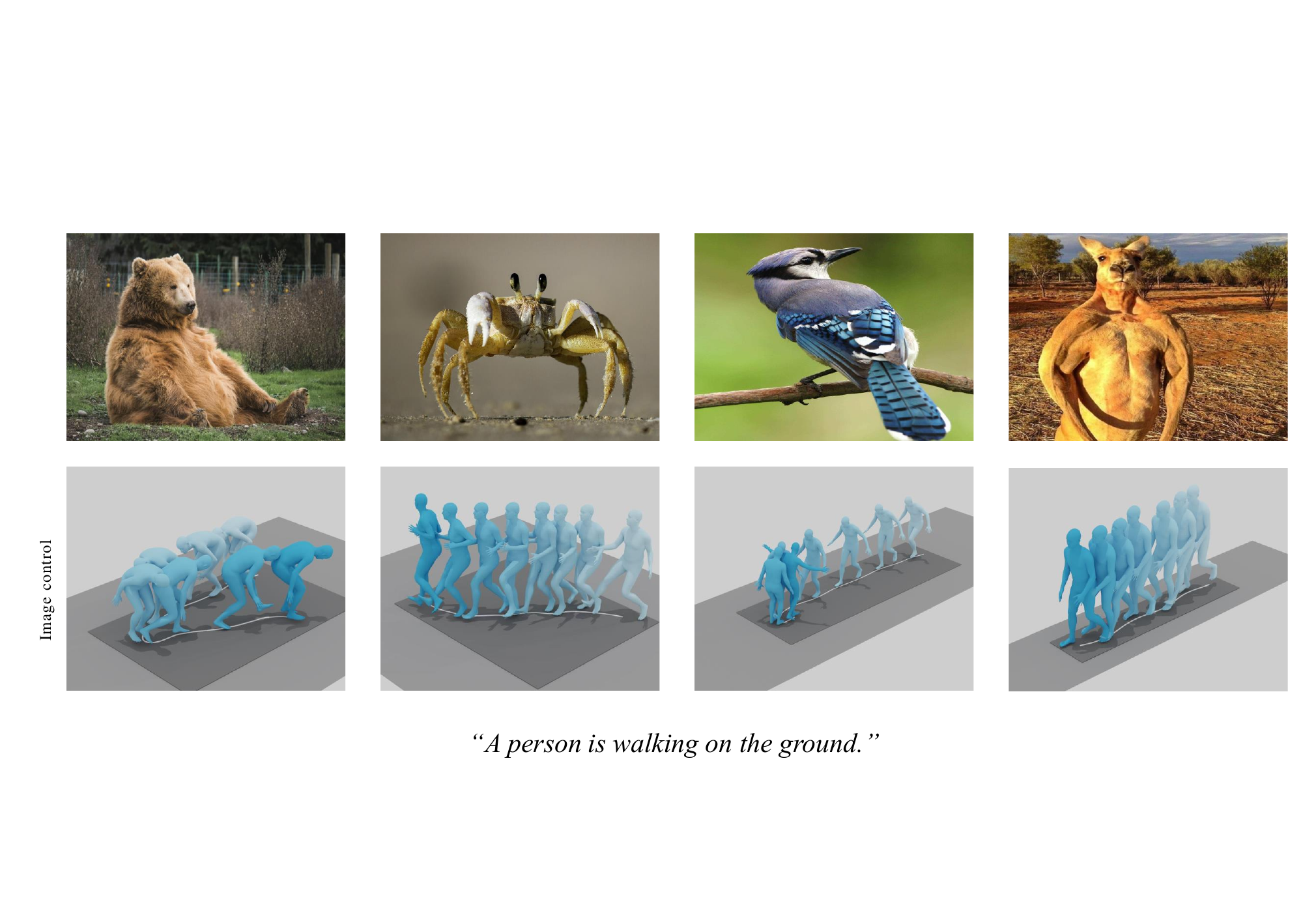} 
	\caption{Image control stylized motion generation of styles not present in the training data, validating the generalization ability of our model.}  
	\label{fig:general} 
\end{figure*}

To further validate the generalization capability of our model in the task of multimodal stylized motion generation, we introduce images corresponding to motion styles not included in the 100STYLE dataset. This experiment is conducted to validate whether the model can infer these unseen motion styles and generate motion sequences that are both consistent with the given content and compatible with the inferred style. As shown in Figure \ref{fig:general}, despite the absence of motion styles such as bear, crab, bird, and kangaroo in the 100STYLE dataset, the model is able to successfully infer and generate corresponding stylized motion sequences. This result robustly demonstrates that the model has a certain level of generalization ability.
Additionally, we have provided a supplemental video, including motion-driven and image/text-driven stylized motion generation.


\end{document}